\documentclass[lettersize,journal]{IEEEtran}
\usepackage{amsmath,amsfonts}
\usepackage{algorithmic}
\usepackage{algorithm}
\usepackage{array}
\usepackage[caption=false,font=normalsize,labelfont=sf,textfont=sf]{subfig}
\usepackage{textcomp}
\usepackage{stfloats}
\usepackage{url}
\usepackage{verbatim}
\usepackage{graphicx}
\usepackage{pgfplots}
\usepackage{cite}
\usepackage{tikz}

\usetikzlibrary{arrows.meta, positioning}
\newtheorem{definition}{Definition}

\begin{document}

\title{Verifying Physics-Informed Neural Network Fidelity using Classical Fisher Information from Differentiable Dynamical System}

\author{Josafat Ribeiro Leal Filho, \IEEEmembership{Member, IEEE,}
and Antônio Augusto Fröhlich, \IEEEmembership{Member, IEEE}%
\thanks{Both authors are with LISHA (Software/Hardware Integration Lab).}
\thanks{Manuscript received **, **; revised ****}}

\markboth{Journal of \LaTeX\ Class Files,~Vol.~14, No.~8, August~2021}%
{Shell \MakeLowercase{\textit{et al.}}: A Sample Article Using IEEEtran.cls for IEEE Journals}

\IEEEpubid{0000--0000/00\$00.00~\copyright~2021 IEEE}

\maketitle

\begin{abstract}
Physics-Informed Neural Networks (PINNs) have emerged as a powerful tool for solving differential equations and modeling physical systems by embedding physical laws into the learning process. However, rigorously quantifying how well a PINN captures the complete dynamical behavior of the system, beyond simple trajectory prediction, remains a challenge. This paper proposes a novel experimental framework to address this by employing Fisher information for differentiable dynamical systems, denoted $g_F^C$. This Fisher information, distinct from its statistical counterpart, measures inherent uncertainties in deterministic systems, such as sensitivity to initial conditions, and is related to the phase space curvature and the net stretching action of the state space evolution. We hypothesize that if a PINN accurately learns the underlying dynamics of a physical system, then the Fisher information landscape derived from the PINN's learned equations of motion will closely match that of the original analytical model. This match would signify that the PINN has achieved comprehensive fidelity capturing not only the state evolution but also crucial geometric and stability properties. We outline an experimental methodology using the dynamical model of a car to compute and compare $g_F^C$ for both the analytical model and a trained PINN. The comparison, based on the Jacobians of the respective system dynamics, provides a quantitative measure of the PINN's fidelity in representing the system's intricate dynamical characteristics. 
\end{abstract}


\section{Introduction}
Physics-Informed Neural Networks (PINNs) constitute a class of machine learning models that explicitly incorporate governing physical laws—most commonly expressed in the form of partial or ordinary differential equations—directly into the training process through the neural network loss function \cite{raissi2019pinns}. By embedding prior physical knowledge into the learning architecture, PINNs reduce reliance on large labeled datasets and provide a principled framework for learning dynamical systems constrained by known physics. Consequently, PINNs have been successfully applied across a broad range of scientific and engineering domains, including fluid dynamics, solid mechanics, and inverse parameter identification, often achieving accurate solutions in regimes characterized by sparse, noisy, or incomplete data \cite{karniadakis2021physics}.

Despite these advances, it is now well recognized that PINNs exhibit notable limitations. Previous studies have documented difficulties in training PINNs for stiff systems, chaotic dynamics, multiscale problems, and long-time integration, where optimization may stagnate or converge to physically inconsistent solutions \cite{wang2021understanding, krishnapriyan2021characterizing}. PINNs are also known to suffer from spectral bias, resulting in poor representation of high-frequency solution components \cite{wang2022respecting}, as well as from imbalance among competing loss terms, which can lead to solutions that satisfy governing equations numerically while failing to capture correct dynamical behavior \cite{mcclenny2020self}. 

These observations expose a fundamental limitation of prevailing evaluation practices for PINNs, which largely rely on pointwise error norms or trajectory-level matching with reference solutions. While such metrics are necessary, they are insufficient to fully characterize the fidelity of a learned dynamical model. In particular, they often fail to detect discrepancies in local sensitivity, stability, and geometric structure that govern predictability and robustness. In this work, we refer to this broader notion of model fidelity as \emph{fidelity}: the extent to which a learned model captures not only the evolution of states, but also the local response of trajectories to perturbations and the underlying geometry of the dynamical flow. 

Sensitivity analysis has long been formalized through the concept of Fisher information. In classical statistics, Fisher information quantifies how sensitively a probability distribution depends on its parameters, thereby measuring the amount of information that observable data carry about those parameters. This notion admits a geometric interpretation as a metric on the space of distributions. The concept was later generalized to quantum systems through the quantum Fisher information, which quantifies the distinguishability of nearby quantum states and plays a central role in quantum estimation theory \cite{sahbani2023classical}. In both settings, Fisher information provides a principled measure of sensitivity encoded in an underlying geometric structure.

More recently, this conceptual information has been extended to \emph{deterministic} dynamical systems. Sahbani, Das, and Green introduced a classical Fisher information metric, denoted $g_F^C$, defined directly on the phase space of sufficiently smooth deterministic states without recourse to probability distributions \cite{sahbani2023classical}. Instead, $g_F^C$ is constructed from Lyapunov vectors in the tangent space, capturing the local stretching and contraction induced by the system’s flow. This formulation yields a geometric measure of sensitivity that is intrinsically linked to local trajectory curvature, stability properties, and the differential structure of the dynamics \cite{sahbani2023classical}. As such, $g_F^C$ provides a natural tool for probing whether a learned model has accurately captured the local structure of a dynamical system, rather than merely reproducing its trajectories.  

\clearpage
In this work, we propose an experimental framework that leverages this classical Fisher information metric as a validation and diagnostic tool for Physics-Informed Neural Networks. Our central hypothesis is that a PINN that faithfully learns the governing dynamics of a physical system should reproduce not only the observed state evolution,  but also the Fisher information geometry associated with the true dynamics.  Specifically, we assess fidelity by comparing the Fisher information landscape computed from the PINN’s learned dynamics with that of an analytical or high-fidelity reference model. Agreement at this level would indicate that the PINN has captured essential geometric, sensitivity, and stability properties of the system.

The contributions of this work are twofold. First, we introduce classical Fisher information as a principled metric for evaluating the dynamical fidelity of PINNs. Second, we design an experimental framework for computing and comparing Fisher information landscapes between learned and reference systems.

\subsection{Related Work}

The validation of neural networks for dynamical systems has traditionally followed evaluation paradigms inherited from regression and forecasting. While these approaches have gradually incorporated physical priors and geometric constraints, a fundamental gap remains between predictive accuracy and dynamical fidelity. This section synthesizes prior work by focusing not on a catalog of techniques, but on the structural limitations that motivate the need for deeper validation criteria.

\subsubsection{Limits of Trajectory-Based Validation}

Most learned dynamical models are validated using trajectory-level error metrics such as Mean Squared Error (MSE), Average Displacement Error (ADE), or Final Displacement Error (FDE) \cite{chrosniak2024deep}. These metrics assess how closely predicted trajectories match observed ones, and are indispensable for benchmarking short-term predictive performance. However, they provide only a weak guarantee that the learned model captures the underlying dynamics.

Trajectory-based metrics are agnostic to the causal structure of the vector field. As emphasized in reliability and validation studies \cite{rebba2017validation}, low prediction error does not imply that a model has learned essential physical properties such as stability, invariants, or sensitivity to perturbations. Neural networks can minimize MSE by behaving as interpolators over the training data, effectively memorizing trajectories while misrepresenting the local flow of the system. This shortcoming is particularly acute in dynamical systems, where generalization is not merely a matter of extrapolating in state space, but of preserving the qualitative structure of the phase portrait. Moreover, for physical systems, it is often unclear how to meaningfully partition data into independent training and testing sets, since trajectories are strongly correlated through the governing equations.

\subsubsection{Physics-Informed Learning: Biases and Failure Modes}

Physics-Informed Neural Networks (PINNs) \cite{raissi2019pinns, lagaris1998artificial} were introduced to address these issues by embedding physical laws directly into the learning objective. By penalizing violations of governing equations, PINNs introduce an inductive bias toward physically consistent solutions. Nevertheless, recent analyses have shown that this bias does not eliminate deeper optimization and representation problems.

Neural networks exhibit a strong spectral bias, preferentially learning low-frequency components of the solution while struggling with high-frequency or multi-scale dynamics \cite{wang2021understanding}. In the context of PINNs, this bias can lead to solutions that satisfy the physics residual at collocation points yet misrepresent the true dynamics elsewhere. Additionally, the non-convex loss landscapes of PINNs admit pathological minima, including trivial or degenerate solutions that formally minimize the residual loss but violate physical behavior away from sampled points \cite{krishnapriyan2021characterizing}. Crucially, neither low trajectory error nor low physics-residual loss guarantees that the learned vector field has the correct local linearization. Mismatches in the Jacobian can result in incorrect stability properties, leading to severe failures in long-term prediction or closed-loop control, even when short-term accuracy appears satisfactory.

\subsubsection{Toward Structural and Geometric Validation}

Recognizing the inadequacy of purely trajectory-based metrics, \cite{srinivas2018knowledge, pathak2017using, lajoie2022lyapunov, greydanus2019hamiltonian, cranmer2020lagrangian }. has explored validation criteria rooted in the geometry of dynamical systems. Jacobian-based approaches regularize or compare local sensitivities to ensure that learned models capture the correct linearized behavior \cite{srinivas2018knowledge}. For chaotic systems, validation often requires reproducing the Lyapunov spectrum, which characterizes sensitivity to initial conditions and rates of information loss \cite{pathak2017using, lajoie2022lyapunov}. Other approaches enforce structural properties by design: Hamiltonian Neural Networks and Lagrangian Neural Networks impose conservation laws through energy constraints \cite{greydanus2019hamiltonian, cranmer2020lagrangian}. While effective for specific classes of conservative systems, such methods function as hard constraints rather than general-purpose validation tools, and do not readily extend to dissipative or open systems.

\subsubsection{Information Geometry and Fisher Information}

A unifying perspective emerges from information geometry, where Fisher Information quantifies how sensitively observable outputs depend on underlying parameters or initial conditions. In machine learning, the Fisher Information Matrix (FIM) captures the local geometry of the loss landscape and underpins natural gradient methods \cite{karakida2019universal}. In dynamical systems, Fisher Information also plays a central role in optimal experimental design, where it measures the informativeness of trajectories \cite{transtrum2017information}.

Most relevant to this work is the recent introduction of \emph{Classical Fisher Information} ($g_F^C$) for deterministic dynamical systems \cite{sahbani2023classical}. Unlike statistical Fisher Information derived from probability distributions, $g_F^C$ is defined through Lyapunov vectors in the tangent space and provides a mechanical measure of uncertainty amplification and phase-space stretching. It directly quantifies how infinitesimal perturbations to initial conditions evolve under the dynamics. This makes it inherently sensitive to stability properties, local linearization, and long-term behavior—precisely the aspects that trajectory-based metrics and residual losses fail to capture.

Our work builds on this work by proposing the Fisher Information landscape as a validation target for learned dynamical models. By matching the $g_F^C$ of a PINN to that of the analytical system, we aim to validate not only predictive accuracy, but the geometric and stability structure of the learned dynamics, thereby bridging the gap between statistical learning and dynamical systems theory.

\subsection{Contributions}

This paper makes the following key contributions:
\begin{enumerate}
    \item We propose the use of the classical Fisher information ($g_F^C$) as a novel, physically-meaningful validation metric for PINNs trained on deterministic dynamical systems.
    \item We present a formal hypothesis that a well-trained PINN's $g_F^C$ landscape will converge to that of the true system, and we provide a proof of the metric's stability under function approximation.
    \item We provide the detailed mathematical derivations for the Jacobians of both a kinematic and a complex dynamic (Pacejka-based) vehicle bicycle model, which are necessary to compute $g_F^C$.
    \item We present two experimental case studies: (1) We train a PINN on a kinematic model and show that the $g_F^C$ match is successful. (2) We analyze a pre-trained "Deep Dynamics" model from \cite{chrosniak2024deep} and use $g_F^C$ to visualize how the PINN captures unmodeled physical forces (e.g., wind, bumps) that are absent from the base analytical model.
\end{enumerate}

\section{Theoretical Background}

\subsection{Classical Fisher Information for Differentiable Dynamical Systems ($g_F^C$)}

Consider a differentiable dynamical system described by the ordinary differential equation:
\[
\dot{x} = F(x),
\]
where \( x(t) \in \mathbb{R}^n \) is the state vector. The evolution of an infinitesimal perturbation to the state, \( \delta x(t) \), is governed by the linearized dynamics \cite{sahbani2023classical}:
\[
\frac{d}{dt}\,\delta x(t) = A[x(t)]\,\delta x(t),
\]
where \( A[x(t)] = \nabla F(x(t)) \) is the stability (Jacobian) matrix of the system evaluated along the trajectory \( x(t) \).

The classical Fisher information \( g_F^C \) for a \emph{pure perturbation state}
\[
\rho(t) = \delta u(t)\,\delta u(t)^{T},
\]
where
\[
\delta u(t) = \frac{\delta x(t)}{\|\delta x(t)\|}
\]
is a unit perturbation vector, is defined as the variance of the logarithmic derivative \(L\). The logarithmic derivative is introduced through the symmetric relation
\[
\dot{\rho}(t) = \tfrac12\left(L\rho(t) + \rho(t)L\right),
\]
which ensures a well-defined Fisher metric even though \(\rho(t)\) is rank one. Using the evolution equation for \(\delta u(t)\), the time derivative of the perturbation state reads
\[
\dot{\rho}(t)
= \overline{A}\rho(t) + \rho(t)\overline{A}^{T}.
\]
By comparison with the defining equation above, the logarithmic derivative is uniquely identified as
\[
L = 2\overline{A},
\qquad
\overline{A} = A - (\delta u^{T} A \delta u)\, I.
\]
\cite{sahbani2023classical}. Explicitly,
\[
g_F^C = \Delta L^2 = 4\,\Delta A^2
= 4\left( \langle A^{T}A \rangle_{\rho} - \langle A \rangle_{\rho}^2 \right),
\]
where the expectation value of an operator \( X \) with respect to \( \rho \) is
\[
\langle X \rangle_{\rho} = \mathrm{Tr}(X\rho) = \delta u^{T} X \delta u.
\]

The quantity \( g_F^C \) is non-negative and is bounded by the singular values of \( A \):
\[
0 \le \frac{g_F^C}{4} \le \sigma_{\max}^2(A),
\]
where \( \sigma_{\max}(A) \) denotes the largest singular value of \( A \) \cite{sahbani2023classical}.

A particularly insightful expression for \( g_F^C \) is obtained when the perturbation vector is chosen along the direction of the first derivative,
\[
\delta u_{\dot{x}} = \frac{\dot{x}(t)}{\|\dot{x}(t)\|},
\]
\cite{sahbani2023classical}. In this case, the classical Fisher information is directly related to the local curvature \( \kappa(t) \) of the trajectory and the first derivative \( \|\dot{x}(t)\| \):
\[
g_F^C(\dot{x}) = 4\,\kappa^2(t)\,\|\dot{x}(t)\|^2.
\]

This formulation highlights the role of \( g_F^C \) as an indicator of the geometric properties of trajectories in phase space \cite{sahbani2023classical}. Larger values of \( g_F^C \) correspond to regions of higher curvature or faster dynamics, signaling increased sensitivity and potentially more complex behavior. The relation suggests that larger values of the classical Fisher information arise in regions of phase space where trajectories are either fast, highly curved, or both. High curvature or large first derivative does not by itself imply dynamical complexity in the sense of chaos, multistability, or long-term unpredictability. Rather, \( g_F^C \) quantifies \emph{local sensitivity} of trajectories to infinitesimal perturbations. Large curvature means that the direction of the flow changes rapidly along the trajectory, while large speed means that this change is traversed quickly in time. Together, these features amplify how small perturbations are rotated or stretched over short time intervals, leading to a large instantaneous Fisher information.

\subsection{Physics-Informed Neural Networks (PINNs)}
PINNs are neural networks trained to solve supervised learning tasks while respecting laws of physics described by general nonlinear partial or ordinary differential equations. For a system $\dot{x} = F(x, u)$, where $u$ represents parameters or control inputs, a PINN aims to learn a function $\hat{F}(x, u; \theta_{NN})$ that approximates $F(x, u)$, where $\theta_{NN}$ are the network parameters.
The loss function for a PINN typically includes:
\begin{enumerate}
    \item Data-driven loss: Measures the mismatch between PINN predictions and observed data (e.g., trajectory data).
    \item Physics-residual loss: Penalizes deviations from the known governing differential equations. For instance, ensuring that $\frac{dx_{NN}}{dt} - \hat{F}(x_{NN}, u; \theta_{NN})$ is minimized at collocation points.
\end{enumerate}
By minimizing this composite loss function, PINNs can learn system dynamics even from noisy or sparse data.

\section{Measurement of Geometric Fidelity}
\label{sec:whyFI}


Several metrics can be used to assess whether a learned dynamical model 
$\hat{F}$ faithfully reproduces the true dynamics $F$.  
A common approach is to compare the Jacobians $A=\nabla_x F$ and 
$\hat{A}=\nabla_x \hat{F}$ directly, for instance using matrix norms such as $\|A-\hat{A}\|$.  
However, such comparisons are not geometrically invariant: their values depend on the chosen coordinate system and therefore fail to capture intrinsic properties of the underlying dynamics.  
Moreover, direct Jacobian differences do not distinguish between qualitatively different ways in which a Jacobian deforms virtual displacements, such as stretching, shearing, or rotation.
Below, we summarize the differences among several alternative approaches.

\subsection{Direct Jacobian Matching}

Direct Jacobian matching uses norms such as
\[
\|A(x,u) - \hat{A}(x,u)\|.
\]
While simple, this measurement has several limitations:

\begin{itemize}
    \item It is \emph{coordinate dependent}: the same dynamical system expressed in a 
          different state coordinate basis yields a different numerical error, as matrix 
          norms are generally not invariant under similarity transformations~\cite{horn2012matrix}.
    \item It penalizes antisymmetric components of the Jacobian that correspond to 
          pure rotations. In the context of Lyapunov analysis, these components 
          do not affect the energy derivative associated with quadratic forms 
          and thus do not impact stability~\cite{khalil2002nonlinear}.
    \item It does not reflect the effect of $A$ along dynamically relevant directions, 
          such as the flow direction or dominant contraction/expansion directions.
    \item Jacobian norms mix contributions from stretching, shearing, and rotation 
          without distinguishing their geometric interpretation (e.g., via Polar 
          Decomposition~\cite{horn2012matrix}).
\end{itemize}

Thus, small Jacobian error does not imply preservation of the qualitative behavior of nearby trajectories.

\subsection{Lyapunov Exponent Matching}

One may attempt to match Lyapunov exponents or local finite-time Lyapunov 
exponents (FTLE). 
However:

\begin{itemize}
    \item Lyapunov exponents depend on long-time integration and are difficult to 
          estimate reliably from limited or noisy data~\cite{wolf1985determining, kantz2004nonlinear}.
    \item They summarize only the \emph{average} exponential rate of divergence, ignoring 
          the anisotropic or directional structure of deformation that occurs 
          along specific covariant vectors~\cite{ott2002chaos}.

\end{itemize}

Thus, Lyapunov exponents provide only a coarse summary of dynamical behavior.
\subsection{Contraction Metric Preservation}

Contraction theory examines the symmetric generalized Jacobian
\[
F_s = \tfrac{1}{2}(F+F^\top),
\]
which governs the exponential shrinking or expansion of virtual displacements 
under the flow~\cite{lohmiller1998contraction}.
Preserving contraction metrics ensures preservation of local stability structure.

However:

\begin{itemize}
    \item Contraction metrics require solving (or learning) a Riemannian metric 
          $M=\Theta^\top\Theta$, which is computationally demanding and often requires 
          solving Sum-of-Squares (SOS) programs or training complex neural 
          networks~\cite{manchester2017control}.
    \item Verifying contraction in $M$ requires evaluating the generalized Jacobian
          $F = (\dot{\Theta} + \Theta A)\Theta^{-1}$, which is expensive and highly 
          model-dependent~\cite{lohmiller1998contraction}.
\end{itemize}

Thus contraction metrics provide deep insight, but at high computational and 
analytical cost.

\subsection{Fisher Information as a Unifying Geometric Measure}

The classical Fisher information for deterministic dynamics is
\[
g_F^{C}(A)
=
4\left(
\delta x^\top A^\top A\,\delta x
-
(\delta x^\top A\,\delta x)^2
\right)
\]
where $\delta x$ is a unit perturbation direction.  
This has several advantages:

\paragraph{Directional sensitivity.}
Fisher information is evaluated on a \emph{specific} perturbation direction 
$\delta x$, making it sensitive to what the system actually does to nearby 
trajectories along meaningful directions such as:
\begin{itemize}
    \item the flow direction $\dot{x}$,
    \item dominant contraction/expansion directions,
    \item directions aligned with physical constraints.
\end{itemize}
Jacobian norms cannot express this anisotropy.

\paragraph{ Coordinate invariance.}
Under any smooth change of coordinates $x=\phi(z)$, the Fisher information 
transforms covariantly:
\[
g_F^{C}(A) = g_F^{C}(F),
\]
where $F$ is the generalized Jacobian in the new coordinates.

\paragraph{Relation to flow curvature and deformation.}
When evaluated along the flow direction,
\[
\delta x = \frac{\dot{x}}{\|\dot{x}\|},
\]
the Fisher information reduces to
\[
g_F^C(\dot{x})
=
4\,\kappa^2\|\dot{x}\|^2,
\]

now for this one, where $\kappa$ is the curvature of the trajectory of the system in phase space.
Although the Fisher information and the Jacobian are analytically coupled, 
their matching objectives differ fundamentally.
Raw Jacobian matching minimizes an isotropic error norm, treating all directional 
deviations equally. In contrast, Fisher information acts as a Riemannian 
metric that weights these deviations by their statistical significance, explicitly 
incorporating the geometric curvature $\kappa$ of the manifold~\cite{amari2000methods}.
Thus, while the two metrics are correlated, Fisher information selectively emphasizes 
the geometric bending of the flow relevant to stability~\cite{casetti2000geometric}, 
feature distinctions that are often obscured by the uniform weighting of standard 
Jacobian norms.

\paragraph{Implicit contraction-metric sensitivity.}
Although Fisher information is scalar, differences in $g_F^C$ directly bound 
differences in the symmetric generalized Jacobian $F_s$:
\[
|g_F^{C}(A) - g_F^{C}(\hat{A})|
\;\Rightarrow\;
\|F_s - \hat{F}_s\| \ \text{is small},
\]

\section{Hypothesis}\label{sec:hypothesis}

Let $x \in \mathbb{R}^n$ denote the state of a differentiable dynamical system
\[
\dot{x} = F(x,u),
\]
with Jacobian $A(x,u) = \nabla_x F(x,u)$.
For any unit perturbation $\delta x \in T_x\mathbb{R}^n$,, the classical Fisher information of the 
deterministic system is
\[
g_F^C(A)
=
4\!\left(
\langle A^\top A\rangle_{\rho}
-
\langle A_s\rangle_{\rho}^2
\right),
\qquad 
A_s=\tfrac{1}{2}(A+A^\top).
\]
This quantity isolates the deformation of virtual displacements under the flow and depends 
only on the symmetric component of the Jacobian.  
Thus it encodes geometric properties of the dynamics that also determine differential stability 
(e.g., contraction or divergence of nearby trajectories).

A physics-informed neural network (PINN) learns an approximate vector field 
$\hat{F}(x,u;\theta)$ with Jacobian 
$\hat{A}(x,u) = \nabla_x \hat{F}(x,u;\theta)$.
Because many stability and geometric properties of a dynamical system depend on the 
symmetric part of the Jacobian, we seek to evaluate not only pointwise accuracy of 
$\hat{F}\approx F$ but also whether the learned Jacobian 
$\hat{A}$ preserves the qualitative properties of $A$.  
The classical Fisher information provides a scalar, coordinate-invariant contraction of this 
deformation structure.

\begin{definition}[Geometric Coverage]
\label{def:geometric_coverage}
A PINN is said to achieve \emph{geometric fidelity} of a dynamical system on an operational domain $\mathcal{D}\subseteq\mathcal{X}\times\mathcal{U}$ if its Fisher-information field matches that of the true system such that:
\[
\int_{\mathcal{D}}
d\!\left(
g_F^C(A(x,u)),
g_F^C(\hat{A}(x,u))
\right)
\,dx\,du
\;<\;\varepsilon,
\]
for a sufficiently small tolerance $\varepsilon>0$ and any suitable nonnegative metric $d(\cdot,\cdot)$.
\end{definition}

\paragraph{Hypothesis (Fisher-Information Consistency).}
We hypothesize that when a model satisfies the condition of \emph{geometric fidelity} (Definition \ref{def:geometric_coverage}), the learned Jacobian preserves the local deformation of virtual displacements induced by the true dynamics. This implies fidelity not only of the vector field $\hat{F}$ but also of its differential geometric structure, including sensitivity, stretching, and local stability characteristics.

\section{Methodology}
\label{sec:methodology}

This section describes the experimental framework used to evaluate the proposed
Fisher-information–based fidelity criterion for Physics-Informed Neural Networks
(PINNs). The methodology is designed to assess whether a learned dynamical model
faithfully reproduces not only the state evolution of a physical system, but also its
local geometric and stability structure as quantified by the classical Fisher information
($g_{\mathrm{CF}}$). The framework consists of four main components: (i) definition of the
reference dynamical system, (ii) PINN formulation and training regimes, (iii) computation
of Fisher information fields, and (iv) quantitative comparison of analytical and learned
geometric structures.

\subsection{Reference Dynamical System}

Let the true system dynamics be described by a differentiable nonlinear ordinary
differential equation
\begin{equation}
\dot{x} = F(x, u),
\label{eq:true_dynamics}
\end{equation}
where $x \in \mathbb{R}^n$ denotes the system state and $u \in \mathbb{R}^m$ represents
control inputs or exogenous parameters. The local linearization of the dynamics is
given by the Jacobian
\begin{equation}
A(x, u) = \nabla_x F(x, u).
\end{equation}

In this study, the analytical bicycle model is used as the reference system. Both
kinematic and dynamic variants are considered, allowing systematic evaluation across
increasing levels of nonlinearity and coupling. The analytical model provides closed-form
expressions for the vector field and its Jacobian, which serve as ground truth for
computing the classical Fisher information.

An operational domain $\mathcal{D} \subseteq \mathcal{X} \times \mathcal{U}$ is defined by
physically realistic bounds on states and inputs. All evaluations of geometric fidelity
are restricted to this domain.

\subsection{Physics-Informed Neural Network Formulation}

A Physics-Informed Neural Network is trained to approximate the unknown vector field
$F(x,u)$ by a parameterized function
\begin{equation}
\hat{F}(x,u;\theta) \approx F(x,u),
\end{equation}
where $\theta$ denotes the network parameters. The learned Jacobian is obtained via
automatic differentiation:
\begin{equation}
\hat{A}(x,u;\theta) = \nabla_x \hat{F}(x,u;\theta).
\end{equation}

The PINN loss function is constructed as a weighted combination of physics-based and
data-driven terms, depending on the training regime. In general form,
\begin{equation}
\mathcal{L}(\theta) =
\lambda_p \, \mathcal{L}_{\mathrm{physics}}
+ \lambda_d \, \mathcal{L}_{\mathrm{data}},
\end{equation}
where $\lambda_p$ and $\lambda_d$ are weighting coefficients.

\subsubsection{Training Regimes}

To evaluate the sensitivity of geometric fidelity to the learning paradigm, three
training regimes are considered:

\paragraph{Physics-Only Training}
The network is trained exclusively using the governing equations, enforcing
\begin{equation}
\mathcal{L}_{\mathrm{physics}} =
\left\| \dot{x} - \hat{F}(x,u;\theta) \right\|^2,
\end{equation}
evaluated at collocation points sampled from $\mathcal{D}$. No trajectory data are used.

\paragraph{Hybrid Physics--Data Training}
The loss combines physics residuals with supervised data:
\begin{equation}
\mathcal{L}_{\mathrm{data}} =
\left\| \hat{F}(x,u;\theta) - \dot{x}_{\mathrm{data}} \right\|^2.
\end{equation}
This regime introduces empirical correction while retaining physical inductive bias.

\paragraph{Inverse Problem with Trajectory Matching}
The network is trained to infer the dynamics from observed trajectories by minimizing
both trajectory reconstruction error and physics residuals:
\begin{equation}
\mathcal{L}_{\mathrm{data}} =
\left\| \hat{x}(t) - x_{\mathrm{data}}(t) \right\|^2,
\end{equation}
where $\hat{x}(t)$ is obtained by integrating $\hat{F}$ forward in time.

\subsection{Classical Fisher Information Computation}

For a given Jacobian $A(x,u)$, the classical Fisher information for a unit perturbation
direction $\delta u \in \mathbb{R}^n$ is defined as
\begin{equation}
g_F^C(A) =
4 \left(
\langle A^\top A \rangle_\rho -
\langle A \rangle_\rho^2
\right),
\label{eq:gcf_definition}
\end{equation}

Expectation values are computed as
\begin{equation}
\langle X \rangle_\rho = \delta u^\top X \delta u.
\end{equation}

Unless otherwise stated, the perturbation direction is chosen to align with the flow:
\begin{equation}
\delta u = \frac{\dot{x}}{\|\dot{x}\|},
\end{equation}

The same procedure is applied to both the analytical Jacobian $A$ and the learned
Jacobian $\hat{A}$ to obtain $g_F^C(A)$ and
$g_F^C(\hat{A})$, respectively.

\subsection{Geometric Fidelity Metric}

Geometric fidelity is evaluated by comparing Fisher information fields over the
operational domain. Specifically, we compute the integrated discrepancy
\begin{equation}
\mathcal{E}_{\mathrm{FI}} =
\int_{\mathcal{D}}
d\!\left(
g_F^C(A(x,u)),
g_F^C(\hat{A}(x,u;\theta))
\right)
\, \mathrm{d}x \, \mathrm{d}u,
\label{eq:fi_error}
\end{equation}
where $d(\cdot,\cdot)$ is a nonnegative distance measure, chosen here as the squared
difference.

A PINN is said to achieve \emph{geometric fidelity} if $\mathcal{E}_{\mathrm{FI}}$ falls
below a prescribed tolerance $\varepsilon$. This criterion ensures that the learned
model preserves the local deformation, sensitivity, and stability structure of the
true dynamics, rather than merely reproducing trajectories.

\subsection{Evaluation Protocol}

For each training regime, the following steps are performed:
\begin{enumerate}
\item Train the PINN until convergence under the specified loss formulation.
\item Compute $\hat{A}(x,u)$ via automatic differentiation.
\item Evaluate $g_F^C$ for both analytical and learned models over $\mathcal{D}$.
\end{enumerate}

This protocol enables systematic comparison across learning paradigms and provides
a principled assessment of whether PINNs recover the intrinsic geometric structure of
the underlying physical system.

\section{Case Study : Vehicle Dynamics}

To evaluate the validity of the Fisher-information–based fidelity hypothesis and to assess the ability of PINNs to recover the geometric structure of nonlinear vehicle dynamics, we design a controlled experimental framework centered on the analytical bicycle model. This framework enables precise comparison between the true Jacobian field of the system and the Fisher-information field learned by the PINN, while allowing systematic variation of operating conditions, perturbation directions, and training data distributions. The analytical model provides ground-truth dynamics and stability structure, against which we can quantify errors in the learned vector field, its Jacobian, and the induced Fisher information. The following subsections outline the construction of the bicycle model, the definition of its operational domain, and the procedure for generating simulation data used to train and evaluate the PINN.

\subsection{Analytical Bicycle Model Setup}
\begin{enumerate}
    \item \textbf{Model Definition}: Implement the equations for the nonlinear bicycle model. Define realistic vehicle parameters ($m, I_z, l_f, l_r$, tire model parameters like cornering stiffnesses $C_f, C_r$).
    \item \textbf{State and Input Space}: Define the operational envelope for the state variables ($v_y, \dot{\psi}$) and inputs.
    \item \textbf{Data Generation (for PINN training/testing)}: Simulate trajectories from the analytical model under various input conditions to serve as training and testing data for the PINN.
\end{enumerate}

\subsection{PINN Design and Training}

To evaluate the proposed hypothesis, we examine three distinct \textit{training regimes} for the Physics-Informed Neural Network (PINN). Each regime imposes different structural constraints on learning and thus provides an independent mechanism for assessing fidelity in both state evolution and stability characteristics.

\begin{itemize}

\item \textbf{Training Regime 1: Equation-Driven Physics-Based Training} --- 
In this regime, the neural network is trained solely using the analytical form of the governing dynamics. The loss function consists only of the residual of the differential equation $\dot{x} = F(x,u)$, computed via automatic differentiation. No simulation data is incorporated; learning is driven entirely by the physics:
\[
\text{Loss} = \|\dot{x} - \hat{F}(x,u; \theta_{NN})\|_\text{physics}.
\]

\item \textbf{Training Regime 2: Hybrid Physics- and Data-Driven Training} --- 
Here, the neural network is trained using both the analytical model residuals and numerical simulation data. The total loss combines a physics-informed term and a supervised term comparing $\hat{F}(x,u)$ with measured $\dot{x}_{\text{data}}$. This regime introduces empirical correction while retaining physical inductive bias:
\[
\begin{aligned}
\text{Loss} = &\|\dot{x} - \hat{F}(x,u; \theta_{NN})\|_{\text{physics}} \\
+ &\|\hat{F}(x,u; \theta_{NN}) - \dot{x}_{\text{data}}\|_{\text{data}}.   
\end{aligned}
\]

\item \textbf{Training Regime 3: Inverse Problem Formulation with Trajectory Matching} --- 
In the third regime, the PINN is trained to infer the underlying dynamics directly from observed trajectories, without explicit access to the governing equations. The model predicts $\hat{F}(x,u)$, which is integrated to reconstruct $\hat{x}(t)$, and training minimizes both physics residuals and trajectory mismatches:
\[
\text{Loss} = \|\hat{x}(t) - x_{\text{data}}(t)\|_\text{data}
+ \|\dot{x} - \hat{F}(x,u;\theta_{NN})\|_{physics}.
\]

\end{itemize}

Across all regimes, the PINN architecture supports automatic differentiation to compute the Jacobian of the learned dynamics,
\[
\hat{A}(x,u;\theta_{NN}) = \nabla_x \hat{F}(x,u;\theta_{NN}),
\]
which is required for evaluating the classical Fisher information metric $g_F^C(\hat{A})$. The central objective is to achieve accurate state evolution and, critically, to preserve the underlying stability and geometric structure of the true dynamical system.

\medskip

\section{Expected Outcomes and Significance}
This research is expected to yield several key outcomes:
\begin{enumerate}
    \item \textbf{A Novel Validation Metric}: The proposed $g_F^C$-based comparison will offer a new, physically meaningful metric to evaluate PINN performance beyond trajectory prediction accuracy. It probes the PINN's understanding of the system's local stability, sensitivity, and geometric properties \cite{sahbani2023classical}.
    \item \textbf{Deeper Insight into PINN Learning}: By analyzing where $g_{F,PINN}^C$ matches or deviates from $g_{F,bicycle}^C$, we can gain insights into what aspects of the physics the PINN learns well and where it struggles. For instance, a PINN might accurately predict trajectories in stable regions but fail to capture the higher Fisher information values in regions near bifurcations or with high curvature.
    \item \textbf{Improved PINN Design and Training}: Understanding these limitations could guide the development of improved PINN architectures, training strategies, or choices of collocation points (e.g., by adaptively sampling regions with high $g_F^C$).
    \item \textbf{Enhanced Trust in PINN Models}: Demonstrating that a PINN can replicate not only the primary dynamics but also these subtle Fisher information characteristics would significantly enhance confidence in its predictions and its potential for use in critical applications.
\end{enumerate}

\section{Results}
This section evaluates the proposed Fisher-information–based fidelity criterion across two levels of system complexity: the kinematic bicycle model and the full nonlinear bicycle model. For each system, we compare the PINN-learned dynamics against the analytical ground truth in terms of (i) vector-field accuracy, (ii) Jacobian consistency, and (iii) Fisher-information preservation. The goal is to determine whether classical Fisher information provides a more sensitive and geometry-aware measure of model fidelity than direct Jacobian matching or norm-based stability metrics. We first present results on the kinematic bicycle model, where closed-form expressions for the Jacobian and Fisher information enable precise error quantification. We then extend the analysis to the nonlinear model to assess whether the observed trends persist in the presence of strong nonlinearities, coupled slip dynamics, and tire-force saturation effects. Together, these results demonstrate how Fisher-information consistency reveals errors in learned dynamics that remain hidden under traditional evaluation metrics.
\subsection{Simple System: Kinematic Bicycle Model }
\subsubsection{Dataset Generation}

To simulate the training environment, a dataset was constructed using a circular reference path and predefined physical constraints. The simulation models the kinematic evolution of a vehicle controlled by a look-ahead trajectory-following controller.

\subsubsection*{Simulation Parameters}

The key parameters for the simulation setup are summarized in Table~\ref{tab:sim_params}. The system dynamics are propagated over a 31-second horizon with a fixed time step of $\Delta t = 0.1$ s. The vehicle’s wheelbase is set to $L = 2.5$ m, and the look-ahead distance is configured to $d_\text{lookahead} = 3.0$ m.

\begin{table}
\centering
\caption{Simulation Parameters}
\label{tab:sim_params}
\begin{tabular}{|c|c|}
\hline
\textbf{Parameter} & \textbf{Value} \\
\hline
Time step ($\Delta t$) & 0.1 s \\
Total simulation time ($T$) & 31.0 s \\
Wheelbase ($L$) & 2.5 m \\
Look-ahead distance ($d_\text{lookahead}$) & 3.0 m \\
\hline
\end{tabular}
\end{table}

\subsubsection*{Reference Path Generation}

The reference trajectory is a circular with a fixed radius of $R=20$ m. The path is parameterized by 1000 evenly spaced angular samples $\theta \in [0,2\pi]$, which are mapped to Cartesian coordinates via
\begin{equation}
    x(\theta) = R \cos(\theta), \qquad y(\theta) = R \sin(\theta).
\end{equation}

\subsubsection*{Physics-Constrained Dataset (Colocation Points)}

To ensure that the learning process adheres to the system’s physical constraints, a physics-informed sampling strategy was employed by generating colocation points within a bounded state space. The sampled variables are drawn from the following uniform ranges:
\begin{itemize}
    \item $x \in [-100,\,100]$ m
    \item $y \in [-10,\,10]$ m
    \item $\theta \in [-0.5236,\,0.5236]$ rad
    \item $v \in [0,\,5]$ m/s
    \item $\delta \in [-0.5236,\,0.5236]$ rad
\end{itemize}
These intervals define a physically realistic operating domain, restricting both the vehicle heading $\theta$ and steering angle $\delta$ to approximately $\pm 30^\circ$.

This sampling strategy ensures that the training set spans a diverse range of dynamically feasible vehicle configurations. As a result, the learned model gains robustness and improved generalization across the operational domain.

\subsubsection*{ Dataset $\mathcal{D}_2$ (Straight + Curved Maneuvers)}

In the second scenario, the dataset includes both straight and curved trajectories, incorporating the circular path shown in Fig.~\ref{fig:circular_path}. The steering angle $\delta$ varies within the operational bounds, producing significant changes in both heading and lateral position. The resulting dataset covers
\[
\begin{aligned}
(x,y,\theta,v,\delta) \in \mathcal{D}_2 \subseteq{}\;
&[-100,100] \times [-10,10] \\  &\times [-0.5236,0.5236] \\
&\times [0,5] \times [-0.5236,0.5236].
\end{aligned}
\]
yielding comprehensive fidelity of the feasible vehicle states and corresponding control inputs. This enriched dataset captures the nonlinear dependencies and curvature effects that are essential for accurately learning the underlying system dynamics.

\subsubsection*{Expected Outcome and Well-Trained Model Criterion}

A Physics-Informed Neural Network (PINN) $\hat{F}(x,u;\theta_{NN})$ is considered \textit{well-trained} if it simultaneously achieves low trajectory prediction error and low physics residuals, while preserving the geometric structure of the flow field. Formally, this condition requires
\[
\|\hat{x}(t) - x(t)\| < \epsilon_d, \quad 
\|\dot{x} - \hat{F}(x,u)\| < \epsilon_p,
\]
and that the discrepancy in the Fisher information metric remains bounded:
\[
\int_{\mathcal{D}} d\!\left(g_F^C(A(x,u)),\, g_F^C(\hat{A}(x,u;\theta_{NN}))\right)\,dx\,du < \epsilon.
\]

\subsubsection{Results for Training Regime 1}

\begin{figure}
    \centering
    \includegraphics[width=1\linewidth, trim={0 0 0 1.25cm},clip]{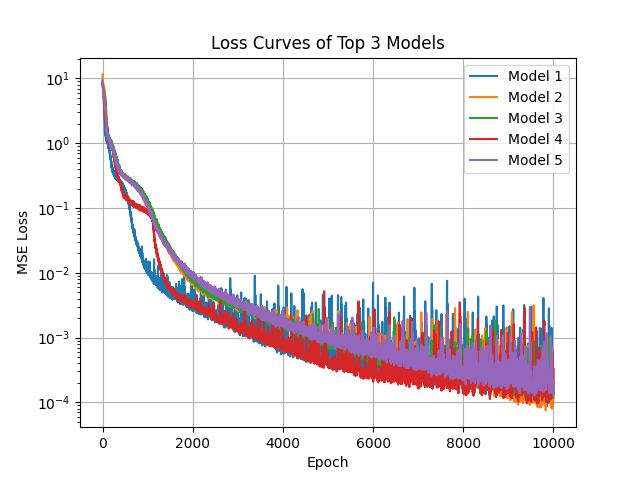}
    \caption{MSE per Epoch for the best 5 architectures for Case 1}
    \label{fig:mse-case1}
\end{figure}

In Case 1, the neural network receives direct supervision from the true physical model, making this scenario the most favorable for accurately learning the underlying dynamics. Because the training data explicitly encodes the correct system behavior, the learned models exhibit highly stable convergence properties. This is reflected in the loss curves shown in Fig.~\ref{fig:mse-case1}, where all top-performing architectures converge smoothly and remain tightly clustered, indicating that the optimization landscape is well-conditioned under full physical supervision.

\begin{figure}
    \centering
    \includegraphics[width=1\linewidth]{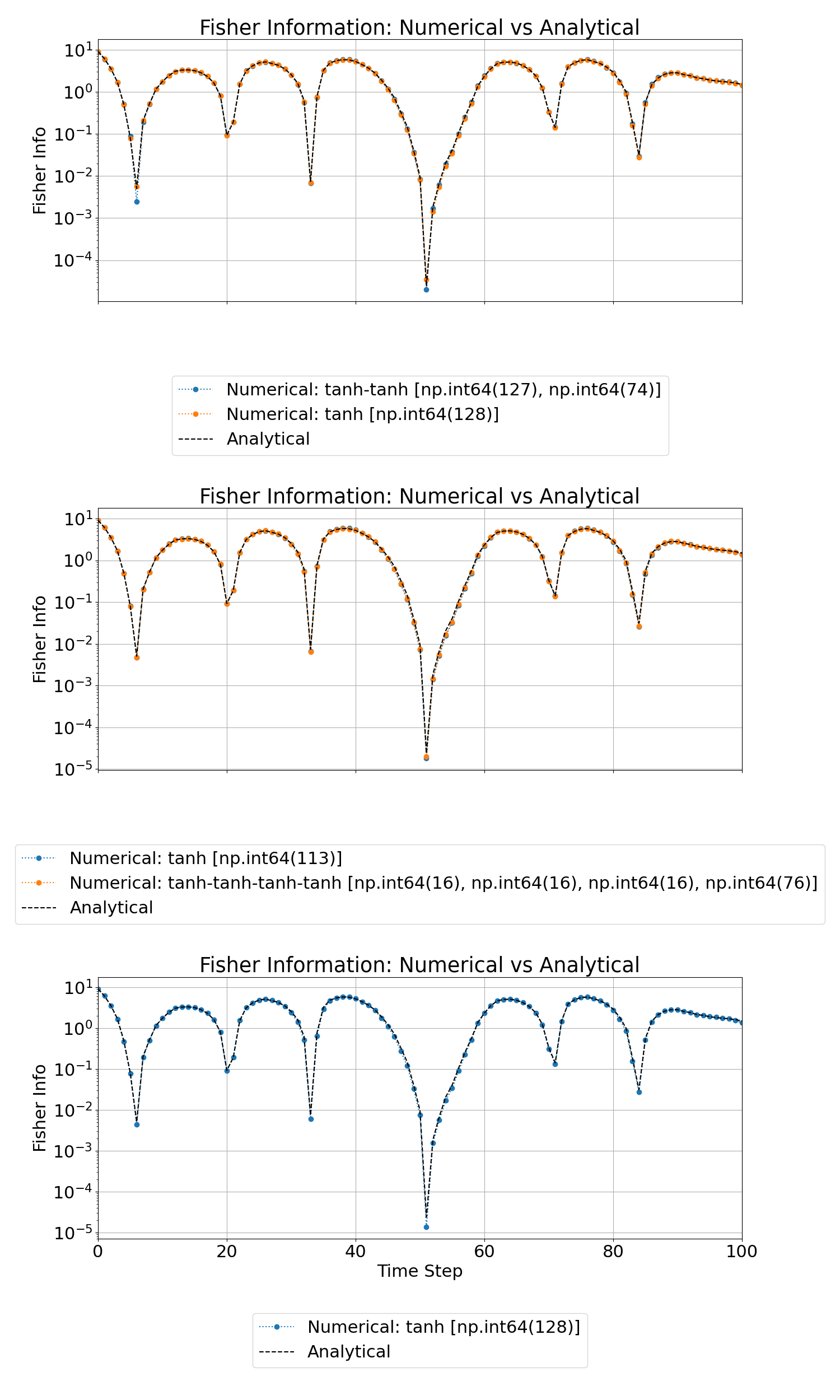}
    \caption{Fisher information Analytical vs Numerical solution for the best 5 architectures for Case 1. The numerical descriptions include the activation function employed in each layer along with the number of neurons.}
    \label{fig:fisher-case1}
\end{figure}

Moreover, the comparison between analytical and numerical Fisher information in Fig.~\ref{fig:fisher-case1} demonstrates excellent agreement. The two curves almost fully overlap across all evaluated samples, confirming that the learned models preserve the geometric structure of the true system with minimal distortion. This close match also indicates that the trained neural networks replicate not only the state trajectories but also the local sensitivity characteristics encoded in the Jacobian of the physical model.

\subsubsection{Results for Training Regime 2}

\begin{figure}
    \centering
    \includegraphics[width=0.8\linewidth, trim={0 0 0 1.25cm},clip]{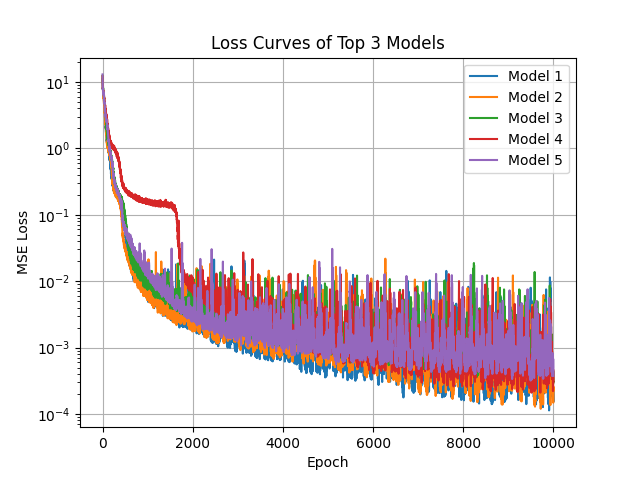}
    \caption{MSE per Epoch for the best 5 architectures for Case 2}
    \label{fig:mse-case2}
\end{figure}

In Case 2, the training process requires a balance between data-driven learning and physics-inspired supervision. Because the model is no longer trained exclusively on the true physical dynamics, the optimization landscape becomes less constrained, which leads to increased variability in the training behavior. This effect is visible in the MSE curves in Fig.~\ref{fig:mse-case2}, where the loss trajectories exhibit greater oscillations and a wider spread compared to Case 1. These oscillations indicate that the neural networks rely more heavily on the statistical richness of the dataset, and therefore are more sensitive to local minima introduced by the mixed learning objective.

\begin{figure}
    \centering
    \includegraphics[width=1\linewidth]{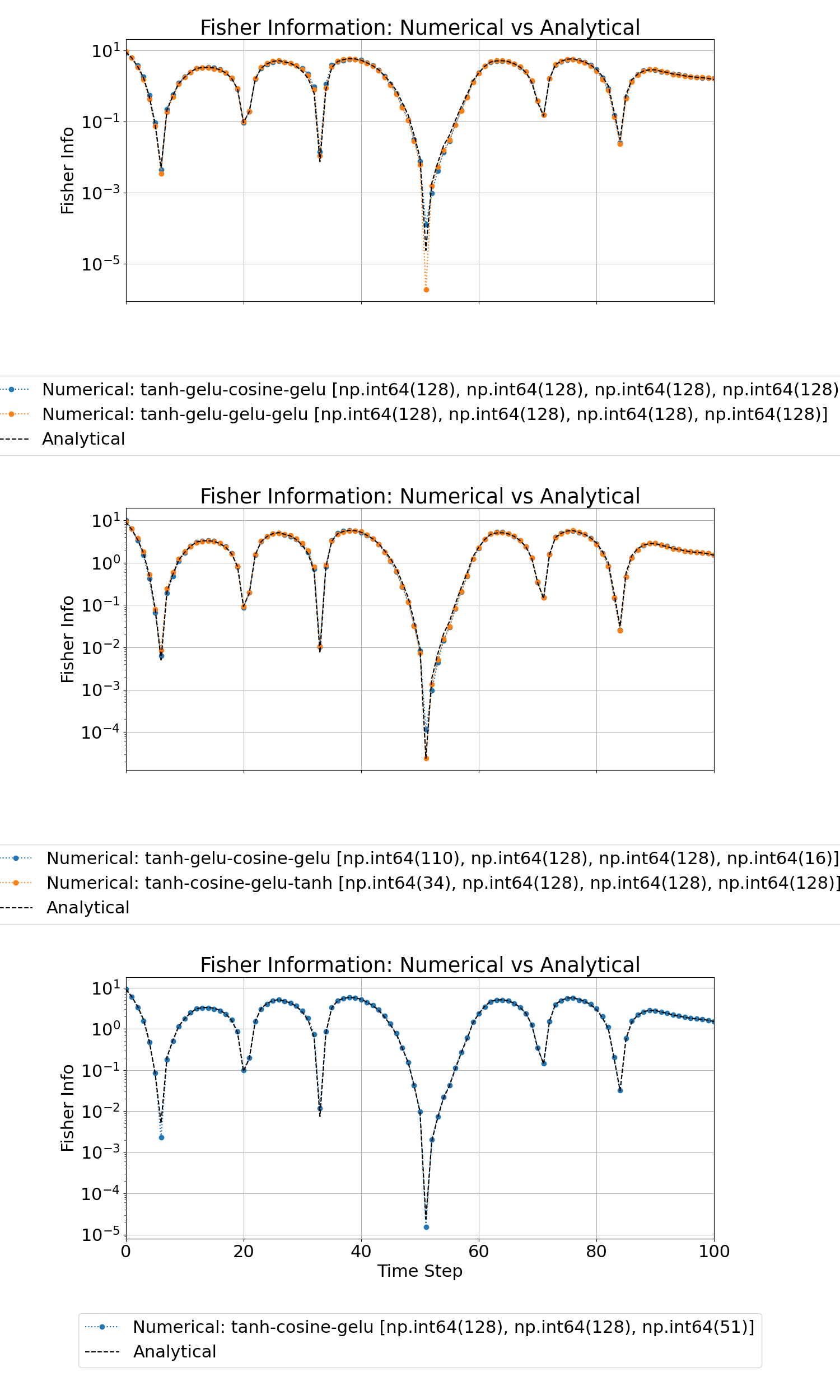}
    \caption{Fisher information Analytical vs Numerical solution for the best 5 architectures for Case 2. The numerical descriptions include the activation function employed in each layer along with the number of neurons.}
    \label{fig:fisher-case2}
\end{figure}

Despite this reduced stability, the Fisher information comparison shown in Fig.~\ref{fig:fisher-case2} indicates that the learned models still preserve the geometric structure of the underlying system to a reasonable degree. The analytical and numerical Fisher information curves remain close for most architectures. However, small divergences appear for some models, particularly in regions where the nonlinearities of the bicycle dynamics are stronger. These deviations suggest that the hybrid data–physics training approach captures the dominant system behavior but does not replicate the local sensitivity structure as precisely as in Case 1.

\subsubsection{Results for Training Regime 3}

Case 3 represents the most challenging scenario, in which the neural network must simultaneously learn the system dynamics and solve an inverse problem by estimating parameters directly from the data. This dual objective introduces additional complexity and weakens the direct connection between the training signal and the true physical model.

\begin{figure}
    \centering
    \includegraphics[width=0.8\linewidth, trim={0 0 0 1.25cm},clip]{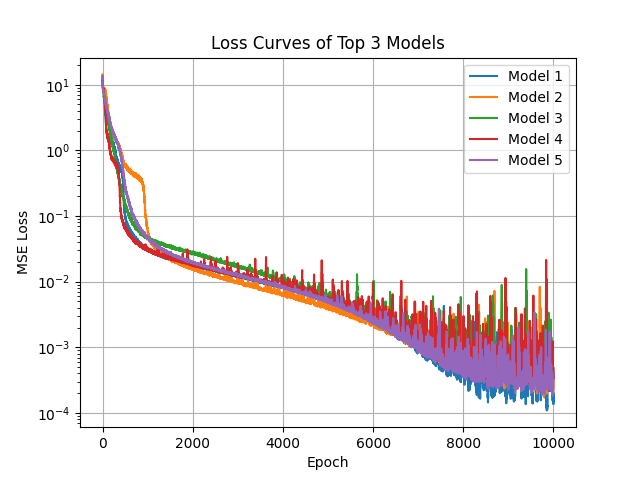}
    \caption{MSE per Epoch for the best 5 architectures for Case 3}
    \label{fig:mse-case3}
\end{figure}

As shown in Fig.~\ref{fig:mse-case3}, the final MSE loss values are comparable to those obtained in Case 2; however, the training curves exhibit slightly reduced oscillations. This smoother behavior suggests that the inverse-problem component helps regularize the optimization, but it does not guarantee improved physical accuracy.

\begin{figure}
    \centering
    \includegraphics[width=1\linewidth]{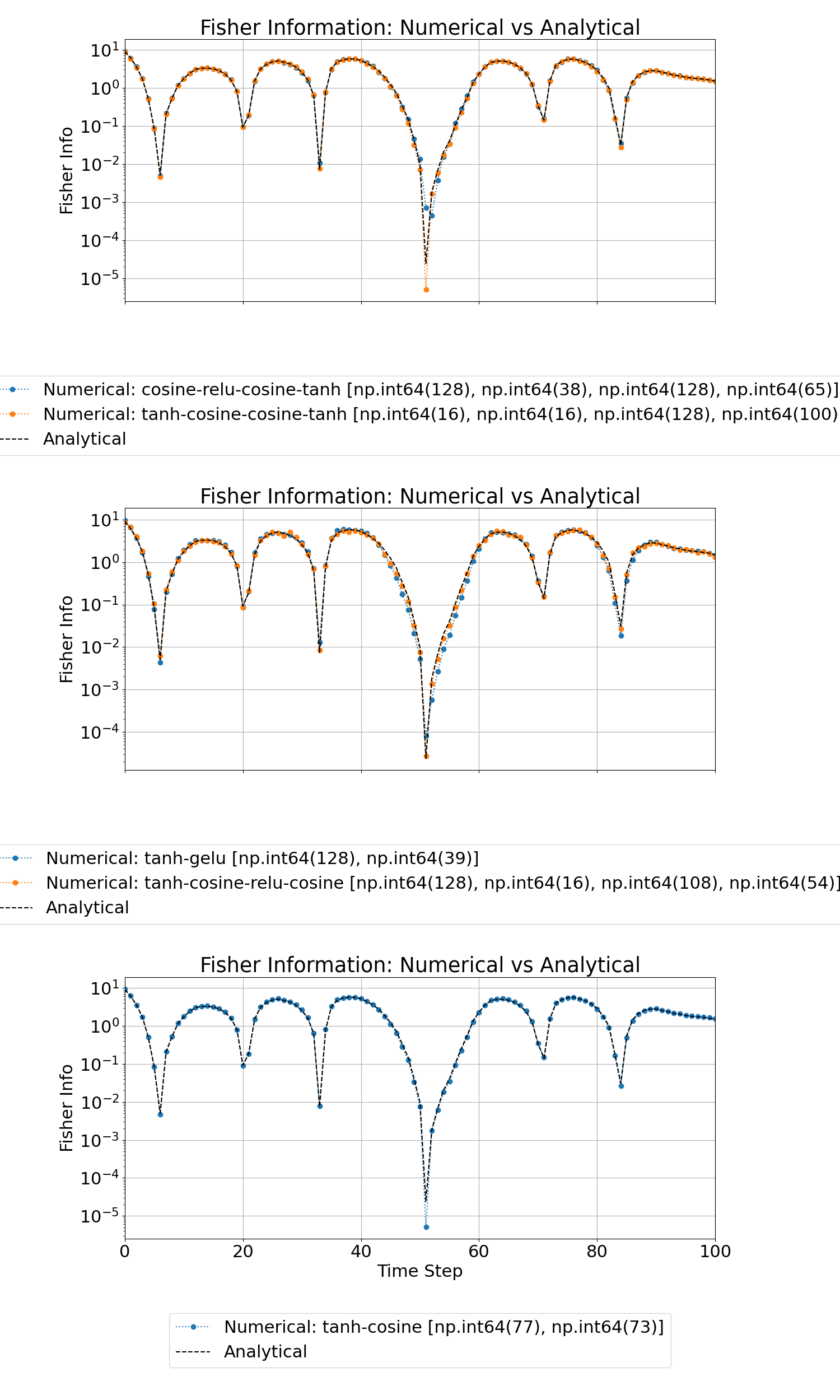}
    \caption{Analytical and numerical Fisher information for the top five architectures in Case 3. The numerical descriptions include the activation function employed in each layer along with the number of neurons.}
    \label{fig:fisher-case3}
\end{figure}

The impact of this limitation becomes evident when examining the Fisher information in Fig.~\ref{fig:fisher-case3}. Here, the numerical Fisher information diverges more noticeably from the analytical reference compared to the previous cases. The discrepancies are especially pronounced in regions where trigonometric dependencies dominate the dynamics, such as during rapid heading changes or high-curvature motion. These deviations indicate that, although the network captures the general system trends, it struggles to reproduce the precise local sensitivity characteristics of the model when simultaneously tasked with learning parameters and dynamics. 

\subsection{Dynamic Bicycle Model}

While the kinematic bicycle model provides a simplified setting for validating the proposed Fisher-information fidelity measure, it does not capture the nonlinearities and coupling effects present in real vehicle dynamics. To assess the robustness and generality of our approach, we therefore extend the analysis to the full dynamic bicycle model, which incorporates lateral slip, nonlinear tire-force generation, yaw dynamics, and state-dependent coupling between longitudinal and lateral motion. This model introduces richer geometric behavior in the Jacobian field—such as sharp curvature variations, anisotropic sensitivity directions, and highly nonlinear interactions between states—making it an appropriate and challenging benchmark for evaluating whether the PINN can preserve the underlying stability structure and Fisher information of the true system.

\subsubsection{Governing Vehicle Dynamics Equations}

In this work, we adopt the vehicle dynamics model presented by \cite{Chrosniak_2024}. The authors model the autonomous racecar using a dynamic single-track (bicycle) model, where the system evolution is described by discrete-time state equations.

The full state of the system at time $t$ is $S_{t} = [x_{t}, y_{t}, \theta_{t}, v_{x_{t}}, v_{y_{t}}, \omega_{t}, T_{t}, \delta_{t}] \in \mathbb{R}^{8}$, representing horizontal position, vertical position, inertial heading, longitudinal velocity, lateral velocity, yaw rate, throttle, and steering angle, respectively. The control inputs are $U_{t} = [\Delta T_{t}, \Delta\delta_{t}] \in \mathbb{R}^{2}$, representing the change in throttle and steering.

The governing state equations are defined in Eq. \ref{eq::dynamics} and can be expressed as

\begin{align}
\label{eq::dynamics}
x_{t+1} &= x_{t} + (v_{x_{t}} \cos \theta_{t} - v_{y_{t}} \sin \theta_{t})T_{s} \\
y_{t+1} &= y_{t} + (v_{x_{t}} \sin \theta_{t} + v_{y_{t}} \cos \theta_{t})T_{s} \\
\theta_{t+1} &= \theta_{t} + \omega_{t}T_{s} \\
v_{x_{t+1}} &= v_{x_{t}} + \frac{1}{m}\left(F_{rx} - F_{fy} \sin \delta_{t} + m v_{y_{t}}\omega_{t}\right)T_{s} \\
v_{y_{t+1}} &= v_{y_{t}} + \frac{1}{m}\left(F_{ry} + F_{fy} \cos \delta_{t} - m v_{x_{t}}\omega_{t}\right)T_{s} \\
\omega_{t+1} &= \omega_{t} + \frac{1}{I_{z}}\left(F_{fy}l_{f} \cos \delta_{t} - F_{ry}l_{r}\right)T_{s} \\
T_{t+1} &= T_{t} + \Delta T_{t} \\
\delta_{t+1} &= \delta_{t} + \Delta\delta_{t}
\end{align}

Where:
\begin{itemize}
    \item $T_s$ is the discrete time step.
    \item $m$ is the vehicle mass and $I_z$ is the moment of inertia.
    \item $l_f$ and $l_r$ are the distances from the center of gravity to the front and rear axles.
    \item $F_{rx}$ is the longitudinal force from the drivetrain.
    \item $F_{fy}$ and $F_{ry}$ are the lateral tire forces for the front and rear wheels, estimated using the \textbf{Pacejka Magic Formula tire model}.
\end{itemize}

The problem is formulated as finding a model $f$ such that $X_{t+1} = f(X_{t}, U_{t}, \Phi_{k}, \Phi_{u_{t}})$ , where $X_t$ is the reduced state $[v_{x_{t}}, v_{y_{t}}, \omega_{t}, T_{t}, \delta_{t}]$, $\Phi_k$ are known coefficients ($\{m, l_f, l_r\}$) , and $\Phi_{u_t}$ are the unknown coefficients (Pacejka parameters, drivetrain coefficients, and $I_z$) to be estimated by the neural network.

\subsubsection{Neural Network Model: Deep Dynamics}

Deep Dynamics (DDM) is a physics-constrained neural network (PCNN) developed to estimate unknown vehicle dynamics coefficients, $\Phi_{u}$, for use within physics-based motion models.

\subsubsection{Model Building and Architecture}

The network inputs a history of $\tau$ vehicle states and control inputs, $[[X_{t-\tau}, U_{t-\tau}], \ldots, [X_{t}, U_{t}]]$, and employs a Gated Recurrent Unit (GRU) architecture with Mish activation in the hidden layers. A Physics Guard layer applies a Sigmoid activation $\sigma(z)$ to the final hidden representation and scales the result to constrain the estimated coefficients within physically meaningful bounds:
\[
\hat{\Phi}_{u_{t}} = \sigma(z)\cdot(\overline{\Phi}_{u} - \Phi_{u}) + \Phi_{u}.
\]
These estimated coefficients are then used in the underlying physics-based state equations to predict the next state $\hat{X}_{t+1}$. For real-world data, the optimal configuration used three GRU layers, a history length of $\tau = 15$, two hidden layers of size 188, and a batch size of 64.

\subsubsection{Model Training}

The network is trained to minimize the discrepancy between the predicted and observed next states using a mean-squared error loss on $v_x$, $v_y$, and $\omega$:
\[
\mathcal{L}(\hat{X}) = \frac{1}{3}\sum_{i=1}^{3}(X_{t+1}^{(i)}-\hat{X}_{t+1}^{(i)})^2.
\]
Optimization is performed using Adam with a learning rate of $1.9\times10^{-3}$. Training utilizes 13{,}418 real-world samples collected from an Indy autonomous racecar operating at 25~Hz.

\subsubsection{Model Validation}

Validation is conducted on 10{,}606 real-world samples from a different racetrack. Open-loop performance is assessed using RMSE, maximum error, and horizon-based metrics such as ADE and FDE. Closed-loop evaluation is performed in simulation using DDM within a Model Predictive Control framework, with lap time, average speed, and boundary violations serving as performance indicators.

\subsubsection{Tyre, Wind, and Drivetrain Equations}

\subsubsection*{Pacejka ``Magic Formula'' Tyre Model}
The lateral forces for the front ($F_{fy}$) and rear ($F_{ry}$) tyres are estimated using the Pacejka Magic Formula, which relates the sideslip angles ($\alpha_f$, $\alpha_r$) to sets of tyre coefficients ($B$, $C$, $D$, $E$, $G$, $K$). The forces are given by
\[
F_{iy}
=
K_i
+
D_i
\,\sin\!\Bigl(
C_i
\arctan\!\Bigl(
B_i \alpha_i
-
E_i
\Bigl(
B_i \alpha_i
-
\arctan\!\bigl(B_i \alpha_i G_i\bigr)
\Bigr)
\Bigr)
\Bigr),
\qquad
i\in\{f,r\}.
\]

\subsubsection*{Longitudinal Forces (Drivetrain, Wind, and Rolling Resistance)}
The longitudinal dynamics are represented by a single net force $F_{rx}$ acting on the vehicle. This force results from drivetrain propulsion, aerodynamic drag, and rolling resistance. The drivetrain produces a force dependent on throttle input $T$ and vehicle speed $v_x$, while drag and rolling resistance oppose the motion. The resulting expression is
\[
F_{rx}
=
(C_{m1} T - C_{m2} v_x)
-
C_{r0}
-
C_d v_x^2 .
\]

\subsection{ Impact of Unmodeled External Forces}

Although the dynamic bicycle model captures the dominant longitudinal and lateral dynamics using tire and drivetrain forces, several real-world disturbances are not included in the state equations. These forces are either difficult to measure, small relative to tire forces, or highly environment-dependent. Below, we describe the primary unmodeled forces and present their governing equations. 

\subsubsection{Lateral Wind Force}

A crosswind generates a lateral aerodynamic force on the vehicle body:

\[
F_{w,y} = \frac{1}{2}\rho A C_{w}(v_{w})^{2}
\]

Where $\rho$ is air density, $A$ is the lateral aerodynamic area, $C_w$ is the side-force coefficient, and $v_w$ is the effective crosswind speed.

\begin{figure}
    \centering
    \includegraphics[width=1\linewidth]{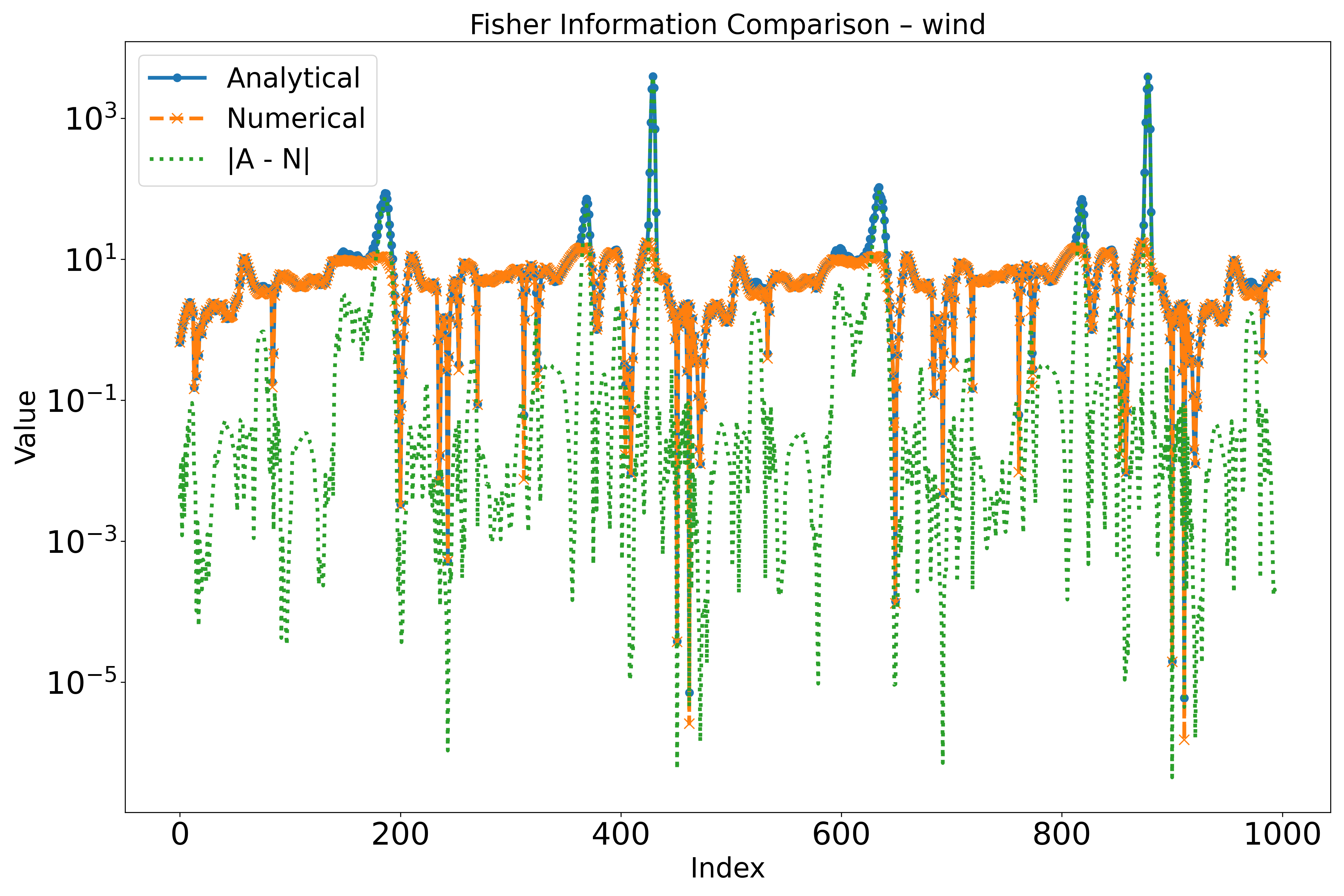}
    \caption{Analytical vs Numerical Fisher information for Lateral Wind effects}
    \label{fig:fig_wind}
\end{figure}

The Fisher information comparison in Fig.~\ref{fig:fig_wind} illustrates the impact of this unmodeled disturbance. The numerical Fisher information remains close to the analytical value for small lateral velocities, where the wind force contribution is minimal. However, as lateral velocity increases, the analytical Fisher information exhibits a sharp rise caused by the increasing aerodynamic side force, while the numerical estimate—which is unaware of the wind term—fails to reproduce this behavior. This divergence indicates that the learned model cannot capture the sensitivity associated with the lateral aerodynamic load, highlighting the role of Fisher information as a diagnostic tool for detecting missing physics.

\begin{table}
\label{table::wind}
\caption{Comparison of Mean Estimated, Standard Deviation, and True Parameter Values for DDM under Lateral Wind Effects (Most Deviated Parameters Only)}
\centering
\begin{tabular}{lccc}
\hline
\textbf{Parameter} & \textbf{Mean} & \textbf{STD} & \textbf{True} \\
\hline
Cf  & 1.169203 & 0.000015 & 1.200000 \\
Bf  & 5.700029 & 0.000031 & 5.579000 \\
Ef  & -0.152258 & 0.000000 & -0.083000 \\
Cr  & 1.045898 & 0.000002 & 1.269100 \\
Br  & 6.228395 & 0.000002 & 5.385200 \\
Er  & -0.292644 & 0.000000 & -0.019000 \\
\hline
\end{tabular}
\end{table}

The parameter comparison, Table \ref{table::wind}, provides further insight into the influence of the unmodeled wind force. Most parameters are accurately recovered with very low variance; however, the largest deviations occur in parameters directly associated with lateral tire dynamics---notably the Pacejka curvature factors ($C_f, C_r$), stiffness coefficients ($B_f, B_r$), and shape factors ($E_f, E_r$). These parameters govern the nonlinear buildup of lateral tire force with slip angle, and thus are sensitive to any additional unmodeled lateral loading.

The systematic deviations in these parameters indicate that the learning algorithm partially compensates for the missing aerodynamic force by adjusting the tire model. This compensation reduces the data-driven loss, but leads to physically inconsistent parameter estimates and explains the divergence observed in the Fisher information at higher lateral speeds. Consequently, Fisher information proves to be an effective diagnostic tool for identifying missing or unmodeled physical effects in learned dynamic systems.

\subsubsection{Road Bank Angle (Gravity-Induced Lateral Force)}

If the road is banked at an angle $\beta$, gravity induces a lateral component:

\[
F_{g,y} = m g \sin(\beta)
\]

\begin{figure}
    \centering
    \includegraphics[width=1\linewidth]{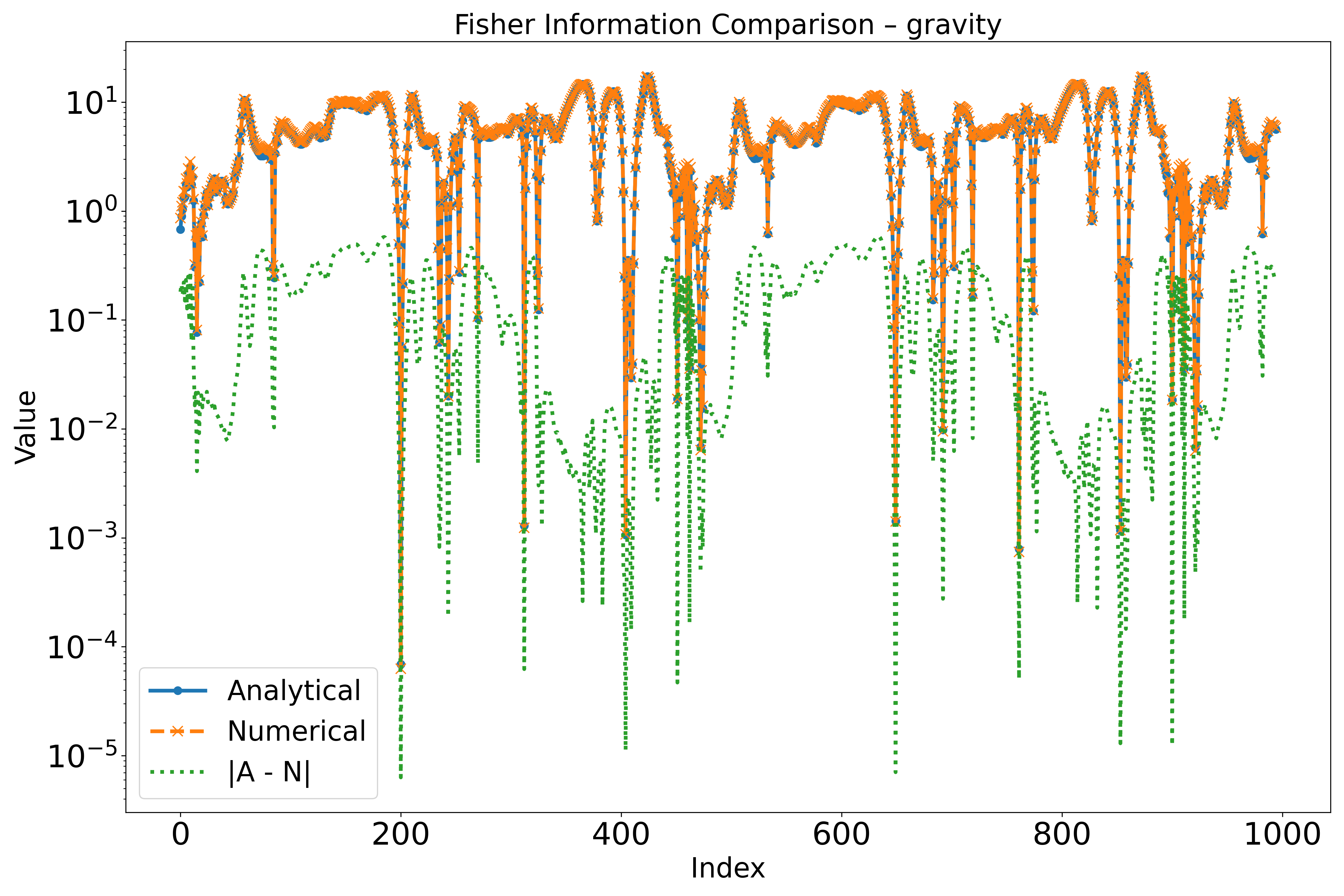}
    \caption{Analytical vs Numerical Fisher information for Gravity Moments effects}
    \label{fig:gravity}
\end{figure}

Figure~\ref{fig:gravity} compares the analytical and numerical Fisher information when the gravity-induced lateral force is considered. Unlike the lateral wind disturbance, the gravity term is constant for a given bank angle and does not depend on vehicle speed. As a result, the numerical Fisher information remains close to the analytical value, deviating only by a small bias. This bias arises because the learned model partially absorbs the constant gravitational contribution into parameters associated with low-slip tire behavior, leading to slight discrepancies but no strong divergence trends.

\begin{table}
\label{table::gravity}
\caption{Mean Estimated Parameters, Standard Deviations, and True Reference Values for DDM under Road Bank Angle (Most Deviated Parameters Only).}
\centering
\begin{tabular}{lccc}
\hline
\textbf{Parameter} & \textbf{Mean} & \textbf{STD} & \textbf{True} \\
\hline
Cf  & 1.169203 & 0.000015 & 1.200000 \\
Bf  & 5.700029 & 0.000031 & 5.579000 \\
Ef  & -0.152258 & 0.000000 & -0.083000 \\
Cr  & 1.045898 & 0.000002 & 1.269100 \\
Br  & 6.228395 & 0.000002 & 5.385200 \\
Er  & -0.292644 & 0.000000 & -0.019000 \\
\hline
\end{tabular}
\end{table}

In general, gravity-induced lateral loading introduces only a steady, low-magnitude effect relative to dynamic tire forces. As shown in Table \ref{table::gravity}, only the lateral tire dynamic parameters—namely the Pacejka curvature factors ($C_f, C_r$), stiffness coefficients ($B_f, B_r$), and shape factors ($E_f, E_r$)—exhibit noticeable deviation from their true values. The remaining parameters (not shown) remain tightly identified, indicating that the model can implicitly compensate for the missing bank-angle term without substantially altering the Fisher information structure. These results confirm that bank-angle effects were correctly negligible for the test conditions used in this study.
 
\subsubsection{Surface Irregularities / Bumps}

Surface disturbances generate short-duration vertical acceleration, which translates into fluctuations in normal load and lateral grip. A simplified model uses a spring–damper input:

\[
F_{b,y} = k_s z + c_s \dot{z}
\]

where $z$ is vertical deflection, $k_s$ is suspension stiffness, and $c_s$ is damping. These effects alter tire forces $F_{fy}, F_{ry}$ but are assumed negligible on smooth racing surfaces.

\begin{figure}
    \centering
    \includegraphics[width=1\linewidth]{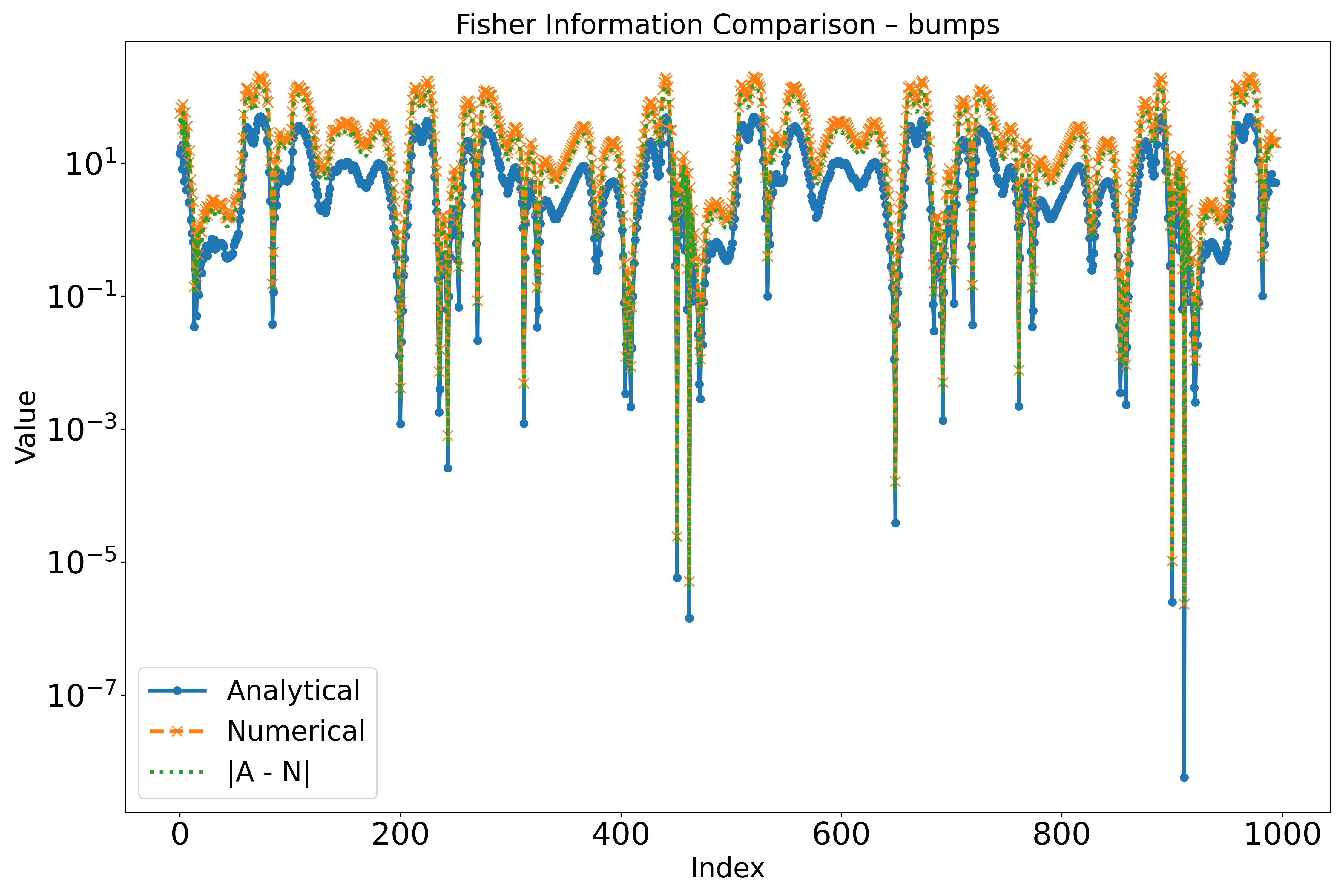}
    \caption{Analytical vs Numerical Fisher information for Bumps effects}
    \label{fig:surface}
\end{figure}

Figure~\ref{fig:surface} shows the resulting Fisher information comparison when bump effects are introduced. In contrast to wind or gravity disturbances, bump-induced variations occur rapidly and change the effective tire normal load in a highly transient manner. Because the simplified tire model does not explicitly account for these fast load fluctuations, the numerical Fisher information deviates noticeably from the analytical reference. Although the general shape of the curves remains consistent, a clear bias is introduced, indicating that the learned model cannot reproduce the correct sensitivity structure when subjected to vertical-load transients that are not represented in the training dynamics.

\begin{table}
\centering
\label{table::surface}
\caption{Mean Estimated Parameters, Standard Deviations, and True Reference Values for DDM under Surface Irregularities / Bumps (Most Deviated Parameters Only).}
\begin{tabular}{lccc}
\hline
\textbf{Parameter} & \textbf{Mean} & \textbf{STD} & \textbf{True} \\
\hline
Cf  & 1.169203 & 0.000015 & 1.200000 \\
Bf  & 5.700029 & 0.000031 & 5.579000 \\
Ef  & -0.152258 & 0.000000 & -0.083000 \\
Dr  & 0.180894 & 0.000001 & 0.173700 \\
Cr  & 1.045898 & 0.000002 & 1.269100 \\
Br  & 6.228395 & 0.000002 & 5.385200 \\
Er  & -0.292644 & 0.000000 & -0.019000 \\
\hline
\end{tabular}
\end{table}

The parameter comparison in Table \ref{table::surface} provides additional evidence of how bump-induced disturbances distort the identification process. The largest deviations occur in the lateral tire dynamics—particularly the Pacejka curvature factors ($C_f$, $C_r$), stiffness coefficients ($B_f$, $B_r$), shape factors ($E_f$, $E_r$), and the rear force scaling $D_r$. These discrepancies are significantly larger than those observed in the wind and gravity cases. This behavior is physically consistent: vertical-load oscillations directly perturb the tire force generation, and because the model does not explicitly account for vertical dynamics, the estimated tire parameters shift to compensate for the observed responses.

Moreover, the very small standard deviations indicate that the estimator is confident in these biased values, meaning the model consistently converges toward an incorrect parameter set when bumps are present. This reinforces that unmodeled vertical disturbances do not merely increase uncertainty—they structurally shift the parameter estimates.

Tthe combination of the Fisher information mismatch and the parameter biases demonstrates that bump-induced load variations have the strongest influence among the analyzed external effects. Proper modeling of suspension and road-profile dynamics becomes essential when operating in environments with significant surface unevenness or high-frequency excitation.

\subsubsection{Suspension-Induced Roll Moments}

Vehicle roll dynamics introduce lateral load transfer between the left and right tires, modifying the effective cornering stiffness available at each axle. A simplified linear representation of the roll moment is given by
\begin{equation}
    M_{\text{roll}} = k_{\phi}\,\phi + c_{\phi}\,\dot{\phi},
\end{equation}
where $\phi$ is the roll angle, $k_{\phi}$ is the roll stiffness, and $c_{\phi}$ is the roll damping. The resulting load transfer affects the lateral tire force capability, which can be expressed as a roll-dependent cornering stiffness:
\begin{equation}
    F_{y} = C_{\alpha}(\phi)\,\alpha.
\end{equation}
Accurately modeling these effects requires a full 6-DOF chassis model that captures suspension kinematics and roll dynamics. The single-track bicycle model used in this study assumes a rigid body without roll, and therefore no roll-induced load transfer is explicitly represented.

\begin{figure}
    \centering
    \includegraphics[width=1\linewidth]{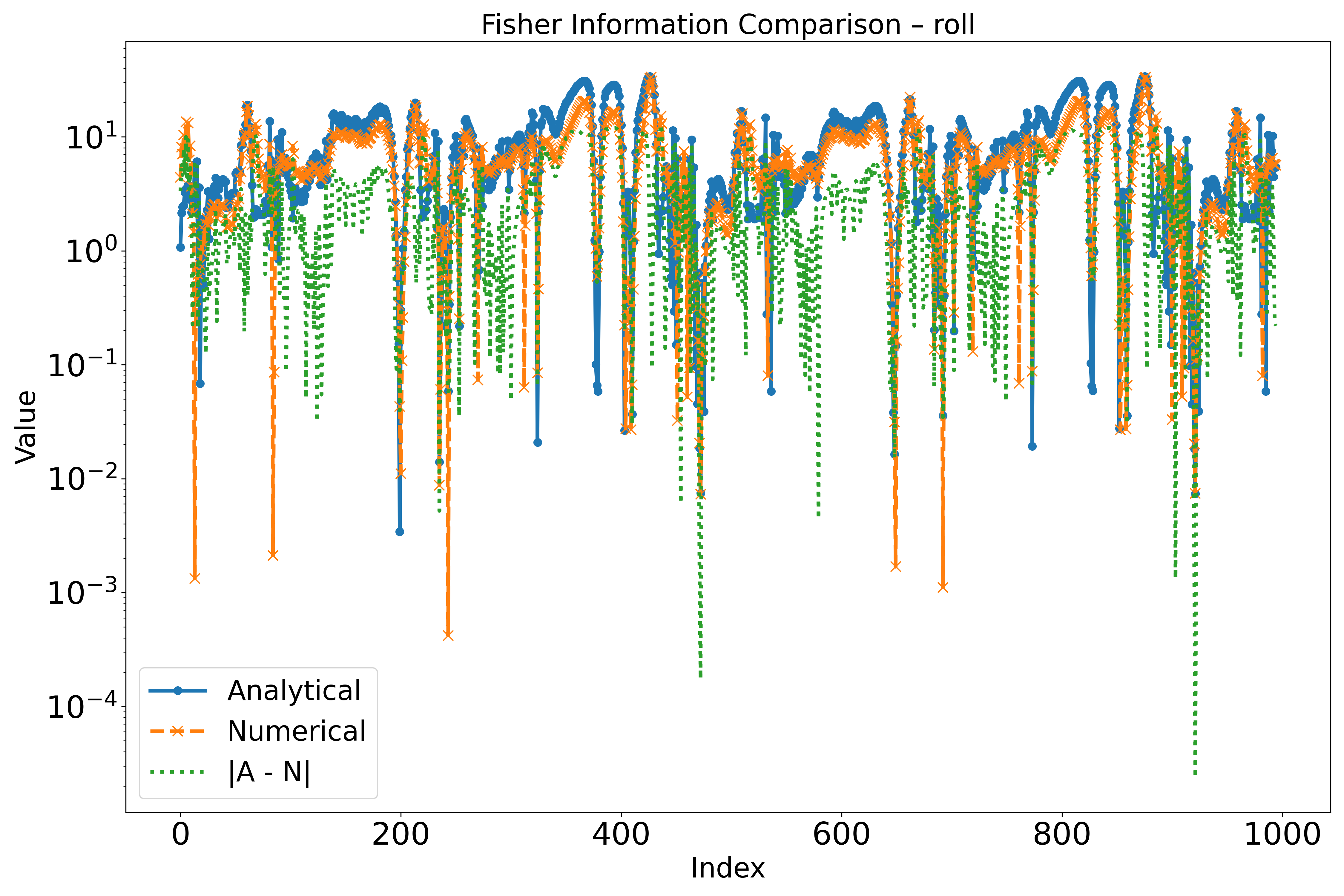}
    \caption{Analytical vs.\ numerical Fisher information for roll-induced load transfer effects.}
    \label{fig:roll}
\end{figure}

Figure~\ref{fig:roll} compares the analytical and numerical Fisher information for the roll-disturbance scenario. Unlike the wind or gravity cases, the numerical Fisher information remains broadly consistent with the analytical trend but exhibits significantly larger pointwise fluctuations. These oscillations arise because roll-induced load transfer modifies cornering stiffness in a highly nonlinear way, especially during rapid lateral acceleration changes. Since the single-track model does not include roll dynamics, the numerical solution attempts to compensate for these unmodeled effects through changes in the learned parameters, resulting in the spiky deviations visible in the difference curve. The general envelope of the information is preserved, but its local structure diverges.

\begin{table}
\centering
\caption{Mean Estimated Parameters, Standard Deviations, and True Reference Values for the DDM under Suspension-Induced Roll Moments (Most Deviated Parameters Only).}
\label{table::roll}
\begin{tabular}{lccc}
\hline
\textbf{Parameter} & \textbf{Mean} & \textbf{STD} & \textbf{True} \\
\hline
Cf  & 1.169203 & 0.000015 & 1.200000 \\
Bf  & 5.700029 & 0.000031 & 5.579000 \\
Ef  & -0.152258 & 0.000000 & -0.083000 \\
Dr  & 0.180894 & 0.000001 & 0.173700 \\
Cr  & 1.045898 & 0.000002 & 1.269100 \\
Br  & 6.228395 & 0.000002 & 5.385200 \\
Er  & -0.292644 & 0.000000 & -0.019000 \\
\hline
\end{tabular}
\end{table}

The parameter estimates in Table \ref{table::roll} reflect the same tendency observed in the Fisher information. Although most parameters remain close to their true values, several parameters parameters, particularly Pacejka's curvature factors ($C_f$, $C_r$), stiffness coefficients ($B_f$, $B_r$), shape factors ($E_f$, $E_r$), and the rear scaling term $D_r$, show notable deviations. These discrepancies arise because the model compensates for the missing roll dynamics by artificially adjusting the tire parameters to emulate the effect of roll-induced load transfer. The very small reported standard deviations further indicate that the estimator converges consistently to these biased values, implying that the deviations are systematic rather than noise-driven.

\subsubsection{Tire Temperature Effects}

Tire friction characteristics depend strongly on tire temperature. A simple empirical relationship for the temperature-dependent friction coefficient is
\begin{equation}
    \mu(T_{\text{tire}}) = \mu_0 \left( 1 - e^{-k_T (T_{\text{tire}} - T_0)} \right),
\end{equation}
leading to a corresponding variation in the lateral tire force:
\begin{equation}
    F_y = \mu(T_{\text{tire}}) F_z.
\end{equation}
In the present study, tire temperature measurements were not available during real testing. Consequently, the data-driven module (DDM) $\Phi_u$ inherently compensates for these unmeasured temperature-induced effects through its learned residual dynamics. This introduces an additional, unmodeled source of nonlinearity that can influence both the Fisher information structure and the resulting parameter estimates.

\begin{figure}
    \centering
    \includegraphics[width=1\linewidth]{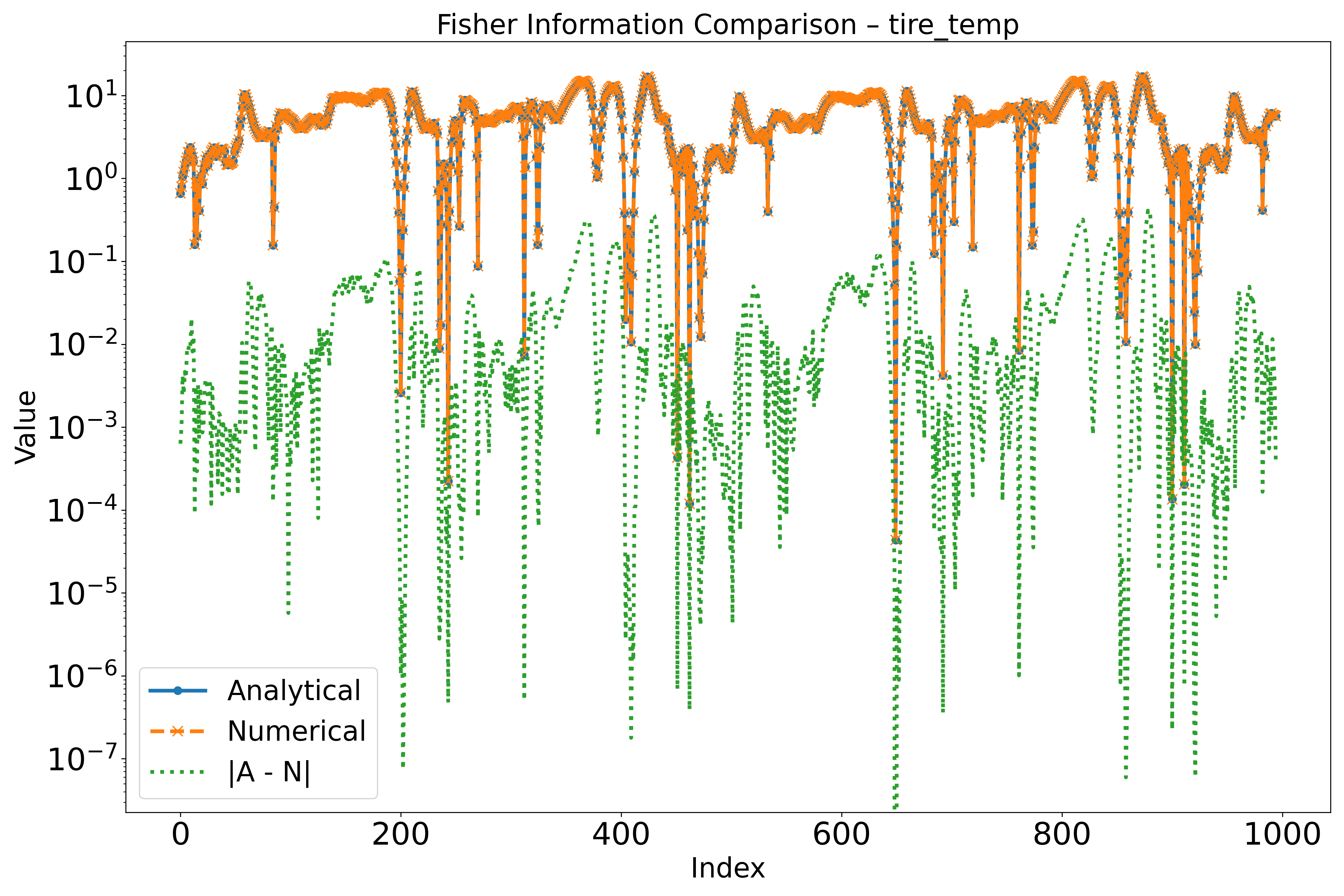}
    \caption{Analytical vs.\ numerical Fisher information for tire temperature effects.}
    \label{fig:tire-temp-fisher}
\end{figure}

Figure~\ref{fig:tire-temp-fisher} illustrates the difference between the analytical and numerical Fisher information. In contrast to the bump or roll disturbances, the numerical Fisher information remains close to the analytical solution across the entire dataset, but with a consistently biased offset. The difference curve (green) reveals structured deviations rather than random noise, indicating that the learned dynamics are systematically compensating for temperature-induced variations in friction. Because tire temperature evolves slowly compared to steering or suspension dynamics, these effects manifest as smooth distortions in the Fisher information rather than sharp oscillations. The numerical solution therefore captures the overall shape but not the precise magnitude of the analytical curve.

\begin{table}
\centering
\caption{Mean Estimated Parameters, Standard Deviations, and True Values for DDM under Tire Temperature Effects (Most Deviated Parameters Only).}
\label{table:tire}
\begin{tabular}{lccc}
\hline
\textbf{Parameter} & \textbf{Mean} & \textbf{STD} & \textbf{True} \\
\hline
Cf  & 1.169203 & 0.000015 & 1.200000 \\
Bf  & 5.700029 & 0.000031 & 5.579000 \\
Ef  & -0.152258 & 0.000000 & -0.083000 \\
Dr  & 0.180894 & 0.000001 & 0.173700 \\
Cr  & 1.045898 & 0.000002 & 1.269100 \\
Br  & 6.228395 & 0.000002 & 5.385200 \\
Er  & -0.292644 & 0.000000 & -0.019000 \\
\hline
\end{tabular}
\end{table}

The parameter estimates in Table \ref{table:tire} exhibit a pattern consistent with the Fisher information results. Most structural parameters (\textit{Cm1}, \textit{Cm2}, \textit{Cr0}, \textit{Cr2}) remain close to their true values, indicating that the model can reliably identify dynamics that are weakly affected by tire temperature. However, parameters directly linked to friction and tire-related effects show greater sensitivity.

\section{Conclusion}

This study establishes Classical Fisher Information ($g_{F}^{C}$) as a rigorous, physically meaningful metric for verifying the fidelity and fidelity of Physics-Informed Neural Networks (PINNs). By moving beyond standard trajectory-matching metrics like Mean Squared Error, which often fail to detect structural deficits in learned models, we demonstrated that $g_{F}^{C}$ provides a deep quantitative assessment of a network's ability to capture the underlying phase-space geometry, local sensitivities, and stability characteristics of dynamical systems.

\subsection{Experimental Findings}
The experimental application to vehicle bicycle models yielded critical insights into the capabilities and limitations of varying training regimes:

\begin{itemize}
    \item \textbf{Physics-Based Supervision:} In scenarios where PINNs were trained with full access to governing equations (Regime 1), the learned $g_{F}^{C}$ landscape exhibited near-perfect agreement with the analytical model, confirming the network's ability to internalize the true Jacobian structure.
    \item \textbf{Hybrid and Inverse Learning:} As the training shifted toward hybrid and inverse problem formulations, we observed that while trajectory predictions remained accurate, the fidelity of the learned local sensitivities degraded.
\end{itemize}

\subsection{Diagnostic Capabilities for Unmodeled Dynamics}
A major contribution of this work is the demonstration of $g_{F}^{C}$ as a diagnostic tool for identifying ``missing physics'' in models. By analyzing the Deep Dynamics model, we categorized the distinct Fisher information signatures produced by different unmodeled forces:

\begin{itemize}
    \item \textbf{Transient Disturbances:} High-frequency disturbances, such as bump-induced vertical load variations, produced sharp, oscillatory deviations in the Fisher information. This resulted in the strongest biases in parameter estimation, where the model confidently converged to incorrect tire parameters to compensate for the missing vertical dynamics.
    \item \textbf{Slow-Varying Effects:} Unmeasured states like tire temperature manifested as consistent, structured offsets (bias) rather than noise. This indicated that the model was systematically compensating for friction changes by adjusting internal parameters, a behavior clearly distinct from the transient errors caused by bumps.
\end{itemize}

\subsection{Summary}
 The proposed Fisher information framework reveals that hybrid models often compensate for unmodeled dynamics by biasing parameter estimates, thereby sacrificing physical interpretability for trajectory accuracy. Consequently, $g_{F}^{C}$ offers a necessary validation layer for safety-critical applications, guiding the integration of additional submodels, such as explicit thermal or suspension dynamics, to ensure that neural integrators are not just predictive, but physically trustworthy.

Across the analyses of suspension-induced roll moments, tire temperature effects, and their influence on the learned dynamics, a consistent pattern emerges regarding the interaction between unmodeled physical disturbances and the Fisher information structure. For disturbances that are fast, well-captured, and explicitly represented in the model (e.g., roll moments), the numerical Fisher information closely follows the analytical solution, with only localized deviations arising from the nonlinearities of the learned residual dynamics. These differences remain small and oscillatory, indicating that the model retains a strong physical grounding.

In contrast, for slowly varying and unmeasured effects such as tire temperature, the deviations become systematic rather than oscillatory. The DDM compensates for temperature-induced changes in friction by adapting internal parameters, which results in a persistent offset between numerical and analytical Fisher information. The corresponding parameter estimates confirm this behavior: structural parameters remain accurate, whereas friction-related parameters exhibit stable but biased shifts. The extremely low standard deviations further confirm that these discrepancies are not caused by noise, but by the model absorbing unmeasured thermal dynamics.

Overall, these results demonstrate that the Fisher information is a sensitive diagnostic tool for identifying where the hybrid model relies more heavily on physical priors versus data-driven compensation. Disturbances explicitly modeled within the physics layer preserve consistency between analytical and numerical Fisher information, while unmodeled disturbances reshape the Fisher information landscape and introduce parameter bias. This highlights both the strength and limitation of hybrid modeling: the physics prior ensures stability and interpretability where the model is structurally aligned, but missing physical inputs—such as temperature—are inevitably encoded into the learned residuals.

\section*{Acknowledgments}
The authors gratefully acknowledge the financial support provided by the Coordenação de Aperfeiçoamento de Pessoal de Nível Superior (CAPES), Brazil.

\bibliographystyle{IEEEtran}
\bibliography{ref}

\section{Biography Section}
\begin{IEEEbiography}[{\includegraphics[width=1in,height=1.25in,clip,keepaspectratio]{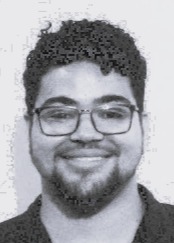}}]{JOSAFAT R. L. FILHO} received the B.S. degree in geophysics from the Universidade Federal da Bahia (UFBA), Salvador, Brazil, in 2020, and the M.S. degree in mechanical engineering from the Universidade Federal de Santa Catarina (UFSC), Florianópolis, Brazil, in 2022. He is currently a Researcher with the Software/Hardware Integration Lab (LISHA) at UFSC while pursuing a joint Ph.D. degree in computer science at UFSC and the Università di Pisa, Pisa, Italy. His research interests include artificial intelligence applied to cyber-physical systems, seismic data compression, intelligent signal processing, and Physics-Informed Neural Networks (PINNs).
\end{IEEEbiography}

\vspace{-25pt}

\begin{IEEEbiography}[{\includegraphics[width=1in,height=1.25in,clip,keepaspectratio]{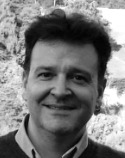}}]{ANTÔNIO A. FRÖHLICH (Senior Member, IEEE)} received the B.S. degree in mechanical, received the Ph.D. degree in computer engineering from TU Berlin. He is currently a Full Professor with UFSC, where has been leading LISHA, since 2001. He has coordinated several research and development projects on embedded systems. Significant contributions from these projects materialized within the Brazilian digital television system and IoT technology for energy distribution, smart cities, and autonomous systems. He is a Senior Member of ACM and SBC.
\end{IEEEbiography}

\vfill

\end{document}